\documentclass{article}
\usepackage{authblk}
\usepackage{subcaption}

\usepackage{graphicx}
\usepackage{booktabs}
\usepackage[accsupp]{axessibility}  

\usepackage{hyperref}

\usepackage{orcidlink}
\usepackage[normalem]{ulem}

\newcommand{\first}[1]{\textbf{#1}}
\newcommand{\second}[1]{\underline{#1}}
\newcommand{\third}[1]{\uuline{#1}}
\def\NuestroMetodo{SV-TAD}
\def\sparseconv{SparseConv2D}
\def\throughputImpr{2.2x}

\def\mAPImprTHUMOS{1.77}

\def\TFLOPSImprBase{43\%}
\def\TFLOPSImprLarge{64\%}

\def\trainSpeedup{1.7x}

\def\TFLOPSRatioInternThu{51\%}
\def\TFLOPSRatioInternAnet{59\%}

\begin{document}

\title{SV-TAD: Native Sparse Convolutions for Efficient Temporal Action Detection}

\author[1]{Ricardo Pizarro\orcidlink{0000-0001-7894-0236}, }
\author[2]{Roberto Valle\orcidlink{0000-0003-1423-1478}}
\author[3]{Jos\'e M. Buenaposada\orcidlink{0000-0002-4308-9653}}
\author[1]{Luis M. Bergasa\orcidlink{0000-0002-0087-3077}}
\author[2]{Luis Baumela\orcidlink{0000-0001-6910-4359}}
\affil[1]{Universidad de Alcal\'a, Alcal\'a de Henares, Spain}
\affil[2]{Universidad Polit\'ecnica de Madrid, Madrid, Spain}
\affil[3]{Universidad Rey Juan Carlos, M\'ostoles, Spain}
\date{ricardo.pizarroc@edu.uah.es}

\maketitle

\begin{abstract}
To adapt billion-parameter Vision Transformers for long-video understanding, recent methods freeze the backbone and train light\-weight convolutional modules.
While effective for parameter-efficient trai\-ning, existing adapters do not reduce inference-time computation, leaving scalability with respect to video length largely unaddressed. Token selection can reduce attention cost by pruning redundant tokens, but it breaks the spatial grid structure required by convolutional adapters. This forces an expensive dense reconstruction, nullifying much of the potential speedup.
We address this by introducing native sparse 2D convolutions, a primitive that allows these adapters, for the first time, to operate directly and efficiently on dynamically pruned token sets.
We integrate this primitive into SV-TAD, an adapter framework for temporal action detection, reducing VideoMAEv2-L computation by up to 64\% and achieving 2.2× faster inference, while maintaining state-of-the-art accuracy on THUMOS-14 and ActivityNet-1.3. When scaled to InternVideoNext-L, our approach surpasses the previous state of the art at roughly half its computational cost. Moreover, the sparse formulation naturally supports auxiliary task tokens, which improves fine-grained assembly detection on ATTACH. Code and trained models are available at \url{https://github.com/pcr-upm/eccv26_tad}.
\\
\\
\textbf{keywords:} Sparse Convolutions, Token Selection, Vision Transformers, Temporal Action Detection, Efficient Inference, Efficient Training
\end{abstract}

\section{Introduction}
\label{sec:intro}
Recent advances in large-scale video foundation models have significantly improved video understanding~\cite{wang2023videomaev2}. To adapt these billion-parameter Vision Transformers to temporal action detection (TAD) tasks without prohibitive fine-tuning cost, adapter-based approaches insert lightweight convolutional modules into a frozen backbone, allowing parameter-efficient specialization~\cite{liuadatad2024, stadapter}. However, despite their parameter efficiency, these adapters still operate on dense token grids and their computational cost scales proportionally with the full sequence length of the video. As video resolution and temporal duration increase, this dense processing becomes a major computational bottleneck. Moreover, existing adapter-based approaches primarily improve training efficiency and do not reduce the inference-time computational cost of the backbone, leaving scalability with respect to video length largely unaddressed.

In parallel, token selection~\cite{liang2022evit, bolya2023tome} has emerged as an effective mechanism to reduce the quadratic cost of self-attention by dynamically pruning uninformative tokens. Unlike architectural modifications, such as efficient attention variants~\cite{tay2022efficient} or distillation~\cite{beyer2022knowledge}, token selection preserves the original model's representations. This makes it uniquely attractive for adapter-based fine-tuning, where any reduction in token count could directly translate into wall-clock savings.

However, while sparsification substantially reduces attention complexity, it disrupts the regular spatial structure required by convolutional adapters, producing a fundamental \textit{representational gap}. This affects convolutions, feature pyramid networks, and any operation that assumes fixed spatial neighborhoods. Existing convolutional adapters reshape tokens into spatial feature maps and apply grid-based convolutions, requiring dense reconstruction when tokens are removed. This reconstruction step substantially diminishes the computational gains of sparsification, so current adapter frameworks cannot fully benefit from token-level sparsity. Additionally, existing 3D sparse convolutional engines~\cite{submanifold, minkowski} cannot be repurposed here, as their format requirements and host-side hash-table operations are architecturally incompatible with the dynamic sparsity of token selection (Sec.~\ref{sec:related}).

\begin{figure}[t]
    \centering
    \includegraphics[width=\linewidth]{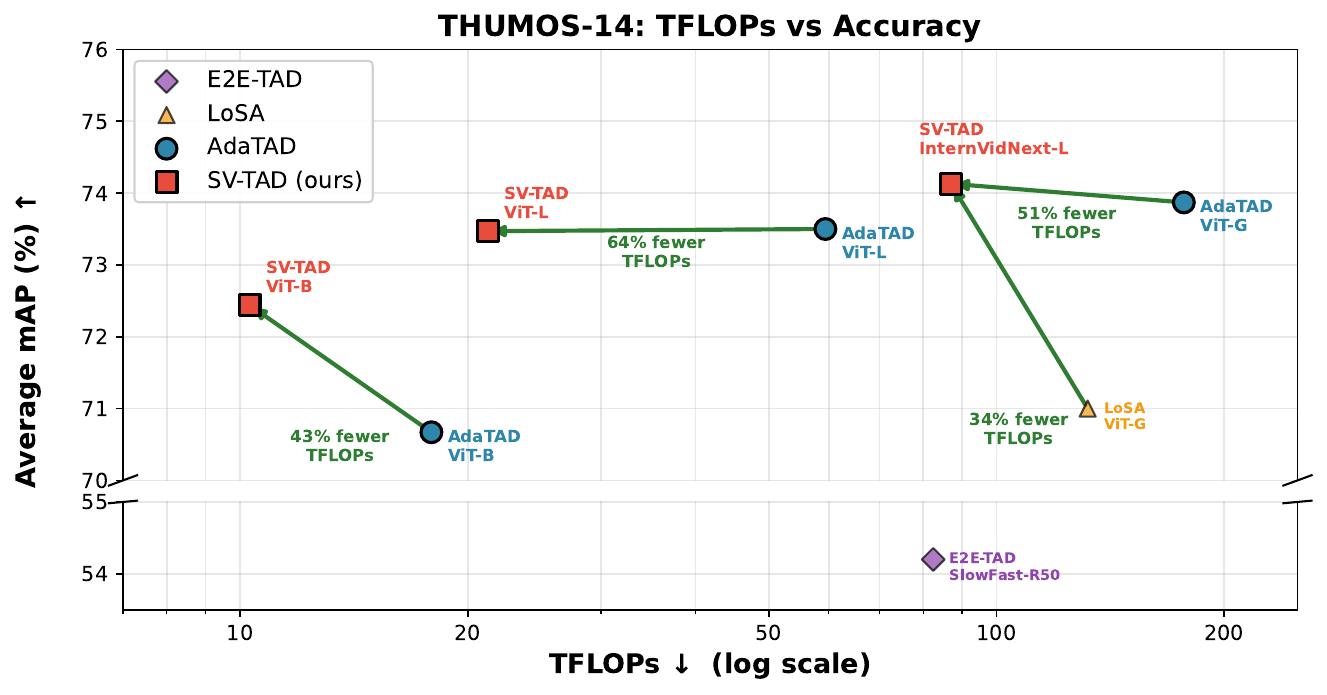}
    \caption{\textbf{TFLOPs vs.\ accuracy on THUMOS-14.} \NuestroMetodo{} (red squares) matches or exceeds AdaTAD~\cite{liuadatad2024} (blue circles) at substantially lower compute across three backbone scales: \textbf{\TFLOPSImprBase{}} fewer TFLOPs with VMAEv2-B, \textbf{\TFLOPSImprLarge{}} with VMAEv2-L, and \textbf{\TFLOPSRatioInternThu{}} when comparing our InternVideoNext-L against AdaTAD's VMAEv2-G.}
    \label{fig:teaser}
\end{figure}

To address this, we introduce \sparseconv{}, a \textit{native sparse 2D convolution} that bridges token selection and convolutions without dense reconstruction. Our key insight is that the sparse token set already contains everything a convolution needs: a convolution does not require the full spatial grid, only the adjacency between each output position and its $K^2$ neighbors; encoding this adjacency explicitly over the sparse set makes the grid unnecessary. We precompute a \textit{neighbor index table} that maps each retained token to its neighbors' positions in the sparse sequence, then execute the convolution through custom CUDA kernels that gather features directly via this table. The regular 2D patch grid further allows the table to be built in $\mathcal{O}(N_{kept})$, with $N_{kept}$ being the number of tokens selected, translating token reduction into proportional speedups in both memory and computation.

We instantiate this primitive in SparseViT-TAD (\NuestroMetodo{}), a temporal action detection framework where sparse convolutional adapters are inserted into each Transformer block of a frozen ViT backbone~\cite{wang2023videomaev2,wang2025internvideo}, while token selection modules progressively prune sequences in strategic layers. As shown in Fig.~\ref{fig:teaser}, \NuestroMetodo{} achieves better efficiency-accuracy trade-offs than existing adapter methods~\cite{liuadatad2024,losa2023}. With ViT-L, it achieves \textbf{\TFLOPSImprLarge{}} fewer TFLOPs and \textbf{\throughputImpr{}} faster throughput.
Our contributions are:
\begin{itemize}
    \item We introduce \textbf{native sparse 2D convolutions} for Vision Transformers, a computational primitive that processes dynamically sparse token sequences through precomputed neighbor index tables and custom CUDA kernels, without dense grid reconstruction.
    \item We design \textbf{sparse adapter architectures} that integrate this primitive into parameter-efficient fine-tuning, enabling convolutional adapters to operate on token-selected sequences for the first time.
    \item We demonstrate \textbf{\TFLOPSImprLarge{}} reduction of TFLOPs, \textbf{\throughputImpr{}} inference speedup, and \textbf{\trainSpeedup{}} training acceleration over the state-of-the-art on THUMOS-14, while maintaining mAP. We validate on ActivityNet-1.3 with consistent gains.
    \item We show that our sparse framework is \textbf{flexible enough to support multitask learning}: in ATTACH, auxiliary landmark supervision improves over the baseline by +2.41\%, demonstrating that secondary tasks can be incorporated with minimal overhead.
    \item We release our CUDA kernels as a \textbf{standalone library} for sparse token processing in PyTorch.
\end{itemize}

\section{Related Work}
\label{sec:related}

\paragraph{Parameter-Efficient Fine-Tuning (PEFT).}
Large-scale ViTs, with hundreds of millions to billions of parameters, cannot be fully fine-tuned for TAD, where each input spans hundreds of frames. Adapter-based methods~\cite{houlsby2019adapter} address this by keeping the backbone weights fixed and learning only small modules inserted between layers.  AIM~\cite{aim} proposes spatial and temporal adapters for video recognition, while ST-Adapter~\cite{stadapter} employs 3D convolutions to capture spatio-temporal patterns.  LoSA~\cite{losa2023} introduces long-short-range adapters and scales to VMAEv2-G, yet underperforms relative to the capacity of its billion-parameter backbone. Other methods such as ViT-TAD~\cite{vit_tad} adapt a short-term ViT-B for clip-level processing. While this keeps per-snippet FLOPs low, ViT-TAD relies on full fine-tuning of all backbone parameters, fundamentally limiting scaling to larger backbones. AdaTAD~\cite{liuadatad2024} applies temporal-informative 1D convolutional adapters, surpassing both LoSA and ViT-TAD while training only ${\sim}$2\% of parameters. A common limitation across these methods is that their convolutional modules always operate on the complete spatial grid, regardless of token informativeness. Recent image-domain work~\cite{kim2025faster,yuan2024efficient} combines PEFT with token selection but requires explicit attention-matrix computation, incompatible with FlashAttention~\cite{dao2022flashattention} and infeasible for TAD sequences exceeding 70k tokens.

\paragraph{Token Selection in Vision Transformers.}
The quadratic complexity of self-atten-tion in the number of tokens $N$ has motivated numerous strategies to alleviate this cost by reducing the number of tokens that traverse the network. EViT~\cite{liang2022evit} ranks patches according to their class-token attention and discards the lowest-scoring ones, while ToMe~\cite{bolya2023tome} progressively merges nearby tokens using bipartite matching, eliminating the need for additional training. DynamicViT~\cite{dynamicvit} instead trains a lightweight decision network that predicts per-token keep/drop masks. In video, TokenLearner~\cite{tokenlearner} and STTS~\cite{stts} extend pruning and merging to the spatio-temporal setting. A shared limitation is that all of these approaches target only the self-attention bottleneck. None specifies how the irregular token sets produced should interact with local operators, such as convolutions. Merging methods carry an additional constraint, as computing the $N{\times}N$ pairwise similarity matrix is at odds with sub-quadratic attention kernels like FlashAttention~\cite{dao2022flashattention}, limiting applicability to long sequences.

\paragraph{Sparse Convolutions.}
Processing spatially sparse data with convolutions has a long history in 3D vision. Submanifold sparse convolutions~\cite{submanifold} and Minkowski-Engine~\cite{minkowski} represent occupied voxels in hash tables, dispatching computation only where data exists, an approach widely adopted for LiDAR point clouds~\cite{spconv}. However, these engines are poorly suited to the dynamic token selection in ViTs: their COO-style sparse-tensor format requires a costly round-trip to the packed dense tensors used by a ViT at every block, and pruning positions invalidates cached kernel maps, forcing the coordinate manager to rebuild all $K^{d}$ maps via hash-table probes per active site with no reuse across layers whose sparsity patterns differ. This hash-based construction is heavier than the problem requires, since ViT coordinates lie on a known regular 2D patch grid, rendering existing 3D sparse engines impractical for adapter-based video models. Separately, sparse-transformer research~\cite{sparsetrans} sparsifies attention patterns rather than convolutional adapters. To our knowledge, native sparse 2D convolutions for the dynamic, layer-varying sparsity of token selection have not been explored before.

\paragraph{Our Proposal.}
Each research line reviewed above addresses one dimension of the efficiency challenge, but leaves the others unresolved. PEFT adapters~\cite{houlsby2019adapter, aim, stadapter, liuadatad2024} always operate on the full spatial grid, so cost scales with total tokens regardless of informativeness. Token selection~\cite{liang2022evit, bolya2023tome, dynamicvit} reduces the count but does not specify how irregular token sets should interact with local operators, limiting savings to attention-only models. Existing sparse engines~\cite{submanifold, minkowski, spconv} are incompatible with ViT tensor formats and unsuitable for dynamic, layer-varying sparsity. \NuestroMetodo{} bridges all three: sparse convolutional adapters provide PEFT-style efficiency, token selection reduces the sequence, and our native sparse 2D convolution enables both to work together, translating token reduction into proportional savings without dense reconstruction or external sparse-tensor engines.

\section{Method}
\label{sec:method}

\begin{figure}[t]
    \centering
        \includegraphics[width=\linewidth]{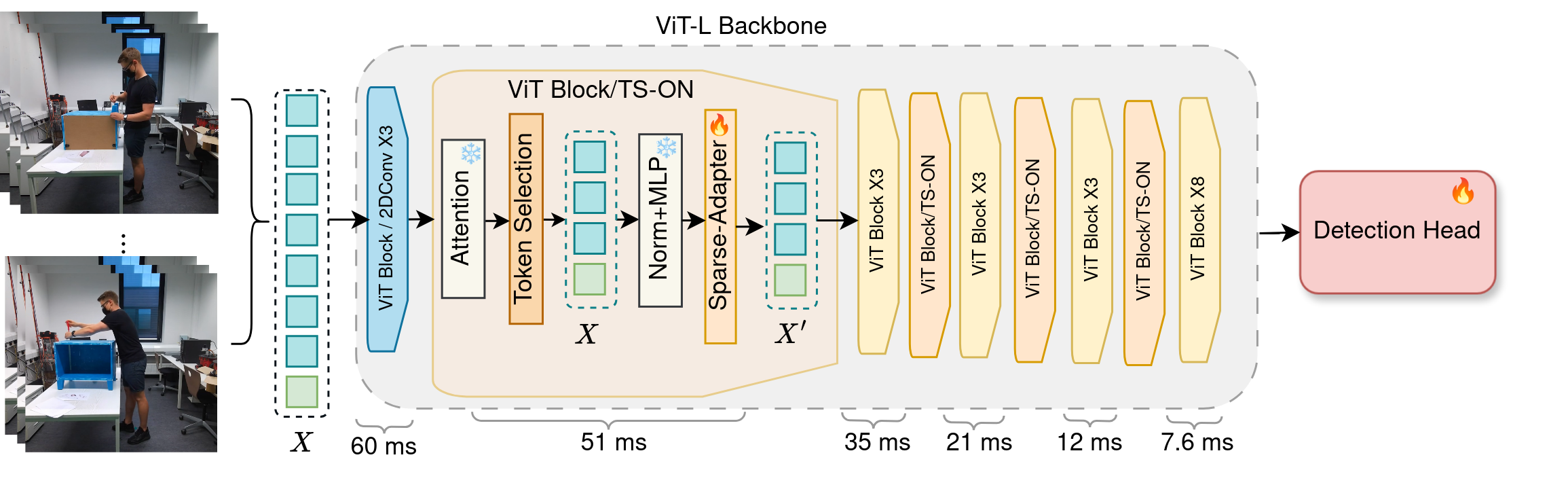}
    \caption{\textbf{\NuestroMetodo{} architecture.} Input frames and auxiliary tokens ($\mathbf{X}$) pass through a frozen pretrained ViT backbone with interleaved token selection (TS-ON) blocks. The first three blocks use standard 2D convolutions, with the rest using \sparseconv{}. The expanded view shows our modified ViT block: frozen attention computes importance scores, token selection prunes uninformative patches, and the trainable sparse adapter operates directly on the sparse sequence. Timing annotations show per-block latency decreasing from 60ms to 7.6ms as the sequence progressively shrinks.}
    \label{fig:method}
\end{figure}

We present \NuestroMetodo{}, a framework that bridges token selection and convolutional adapters for TAD. Central to our approach is \sparseconv{}, a \textit{native sparse 2D convolution} (Sec.~\ref{sec:sparse_conv}), this computational primitive processes dynamically sparse token sequences without dense grid reconstruction. We integrate this primitive into a sparse adapter architecture (Sec.~\ref{sec:adapter}) and show that the resulting framework is flexible enough to support auxiliary task supervision (Sec.~\ref{sec:auxiliary}). Fig.~\ref{fig:method} provides an overview.

\subsection{Token Selection}
\label{sec:token_selection}

We employ attention-based token selection~\cite{liang2022evit, poguise2025} to progressively prune uninformative tokens during the forward pass. Crucially, unlike prior work that discards spatial information after pruning, we \textit{preserve the original grid indices} of kept tokens, this positional information is essential for our sparse convolution.

At designated Transformer blocks, we compute importance scores from aggregated attention weights between \textit{auxiliary tokens} and visual tokens. In the standard setting, only the class token serves as the auxiliary token. When more than one auxiliary task token is present (Sec.~\ref{sec:auxiliary}), importance scores incorporate attention from all $K$ tokens with learnable weights $\mathbf{w} \in \mathbb{R}^{K}$:
\begin{equation}
    \mathbf{s} = \frac{1}{n_h}\sum_{h=1}^{n_h} \sum_{k=1}^{K} w_k \cdot \mathbf{A}^{(h)}_{k}, \quad \mathcal{I}_{kept} = \text{TS}\left(\mathbf{s}, \; \lfloor k_r \cdot N \rfloor\right)
\end{equation}
where $TS$ is the token selection algorithm, $N$ is the number of visual tokens, $\mathbf{A}^{(h)}_{k} \in \mathbb{R}^{N}$ denotes the attention weights from the $k$-th auxiliary token to all visual tokens at head $h$, $n_h$ is the number of attention heads, and $k_r$ is the keep rate. This formulation allows task-specific tokens (e.g.,, auxiliary landmarks) to influence which $N_{kept}$ patches are retained, biasing selection toward regions relevant for the auxiliary task. When $K{=}1$ (class token only), this reduces to the standard EViT formulation~\cite{liang2022evit}. The tokens kept $\mathbf{X}_{vis} = \mathbf{X}_{vis}[\mathcal{I}_{kept}]$ are passed to subsequent layers, while $\mathcal{I}_{kept}$ encodes their original positions $(y, x)$ in the spatial grid $H \times W$.

\paragraph{Compatibility with FlashAttention.}
A critical advantage of our token selection is that it requires only the $K$ attention rows from auxiliary tokens to visual tokens, not the full $N{\times}N$ attention matrix. We extract the required $K$ rows (where $K \ll N$) as a byproduct of the standard attention computation. This allows our entire pipeline, backbone attention, token selection, and sparse adapters to operate with FlashAttention enabled, which is essential for scaling to video sequences.

Following prior work on token selection~\cite{liang2022evit, dynamicvit,poguise2025}, the selection is applied at evenly-spaced layers (blocks 4, 8, 12, 16 for ViT-L), progressively reducing the sequence length. Early layers process all tokens to build robust representations before pruning begins.

\subsection{Native Sparse 2D Convolution}
\label{sec:sparse_conv}

The straightforward way to combine token selection with convolutional adapters is \textit{dense reconstruction}. It consists of scattering the $N_{kept}$ sparse tokens back to their original grid positions $\mathbf{X}_{dense} \in \mathbb{R}^{T \times H \times W \times D}$, filling pruned locations with zeros, applying standard convolution, and then gathering outputs at kept positions. While functionally correct, this approach has fundamental efficiency limitations: (1) memory scales with $\mathcal{O}(THW)$ regardless of sparsity, (2) convolution computes over all $THW$ positions including empty padding, and (3) empirically, speedup plateaus as reconstruction overhead dominates, pruning more tokens yields no additional benefits. The general incompatibility and reconstruction algorithm are explained in detail in Sec.~G of the supplementary material.

\begin{figure}[t]
    \centering
    \includegraphics[width=0.6\linewidth]{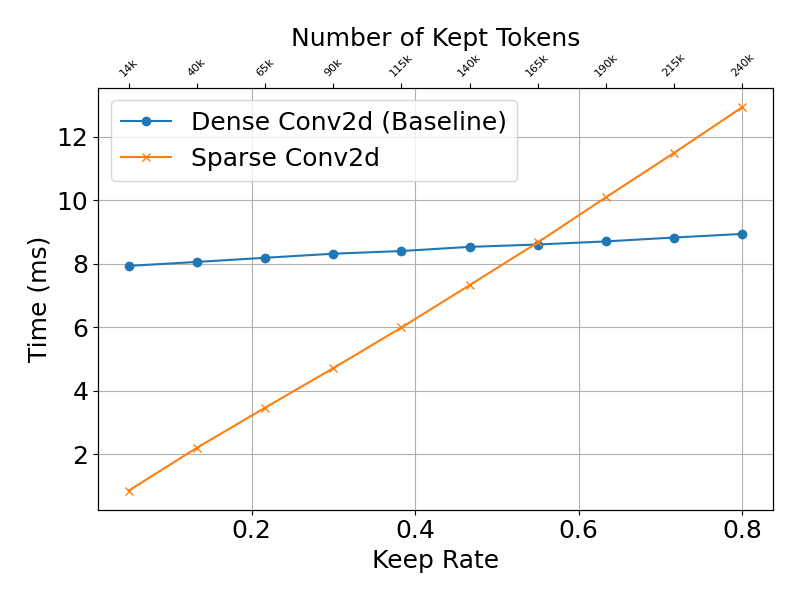}
    \caption{\textbf{Kernel Timing comparison on THUMOS-14.} DenseConv2D (blue) keeps latency nearly constant across different keep rates, while our SparseConv2D (orange) scales linearly with $N_{kept}$.
    }
    \label{fig:sparse_timing}
\end{figure}

We propose a native sparse 2D convolution that operates directly on sparse token sequences, $\mathbf{X_{vis}} \in \mathbb{R}^{N_{kept} \times D/4}$. The key insight is that convolution is inherently \textit{local}, e.g., a $3{\times}3$ kernel only needs 9 neighbors per output, not the entire grid. We exploit this locality through an indirection table that maps sparse tokens to their sparse neighbors, enabling computation without dense reconstruction. For simplicity in the following sections, we use a $3 \times3$ kernel. Fig.~\ref{fig:sparse_timing} shows that our sparse kernel achieves parity with dense convolution at $\sim$55\% keep rate; below this threshold, sparse is strictly faster, reaching $8{\times}$ speedup at 10\% keep rate.

The data structure enabling our sparse convolution is a \textit{neighbor index table} $\mathbf{N} \in \mathbb{Z}^{N_{kept} \times 9}$ that maps each kept token to its spatial neighbors' positions in the \textit{sparse} token list. For kept token $k$ at grid position $(y_k, x_k)$ and kernel offsets $\delta_x \in \{-1, 0, 1\}$ and $\delta_y \in \{-1, 0, 1\}$:
\begin{equation}
    \mathbf{N}[k, j] = \begin{cases}
        \text{sparse\_idx}(y_k + \delta_{y,j}, x_k + \delta_{x,j}) & \text{if neighbor is kept} \\
        -1 & \text{if pruned or out-of-bounds}
    \end{cases}
\end{equation}
where $j \in \{1, \ldots, 9\}$ indexes kernel positions and $-1$ indicates zero-padding.

We construct this table efficiently in two steps, exploiting the regular 2D grid structure:
\begin{enumerate}
    \item \textbf{Build a lookup map.} Create a dense array $\mathbf{M} \in \mathbb{Z}^{H \times W}$ initialized to $-1$. For each kept token $k$ at grid position $(y_k, x_k)$, set $\mathbf{M}[y_k, x_k] = k$. This maps grid coordinates to sparse indices.
    \item \textbf{Fill the neighbor table.} For each kept token $k$, look up its 8 neighbors via $\mathbf{N}[k, j] = \mathbf{M}[y_k + \delta_{y,j}, x_k + \delta_{x,j}]$. If the neighbor was pruned, $\mathbf{M}$ returns $-1$, correctly indicating zero-padding.
\end{enumerate}
Both steps are $\mathcal{O}(N_{kept})$, computed on CPU with negligible overhead (less than $\sim$3ms) relative to the GPU-bound adapter forward pass, and $\mathbf{M}$ is reused across all adapter layers until the next token selection.

\subsubsection{Custom CUDA Kernels.}
We implement forward and backward passes as custom CUDA kernels. The forward pass computes:
\begin{equation}
    \mathbf{y}_k = \sum_{j=1}^{9} \mathbf{W}_j \cdot \mathbf{x}[\mathbf{N}[k, j]] + \mathbf{b}
\end{equation}
where $\mathbf{W}_j \in \mathbb{R}^{C_{out} \times C_{in}}$ is the weight slice for kernel position $j$, and $\mathbf{x}[\mathbf{N}[k, j]]$ gathers neighbor features (zeros if $\mathbf{N}[k, j] = -1$). We provide two strategies, automatically selected based on $C_{out}$: an implicit GEMM path via cuBLAS for large channels ($C_{out} \geq 256$) that fully utilizes Tensor Cores, and custom tiled kernels for smaller channels that fuse gather and GEMM into a single launch. Full kernel design details are in Sec.~A of the supplementary.

\subsubsection{Complexity Analysis.}
Our sparse convolution requires $\mathcal{O}(N_{kept} \cdot K^2 \cdot C_{in} \cdot C_{out})$ operations versus $\mathcal{O}(THW \cdot K^2 \cdot C_{in} \cdot C_{out})$ for dense convolution. With $N_{kept} = k_r \cdot THW$, compute scales linearly with keep rate $k_r$. Memory is decoupled from total sequence length through chunked processing, $\mathcal{O}(\text{chunk\_size} \cdot (C_{in} + C_{out}))$. The full model applies token selection at $S$ stages (e.g.,, $S=4$ at blocks 4, 8, 12, 16 for ViT-L), with cumulative savings compounding across layers. At $k_r{=}0.6$ per stage, the effective rate after two stages drops to $0.36$ and after four to 0.13, placing the majority of layers well below the per-kernel crossover point where sparse convolution outperforms dense (Fig.~\ref{fig:sparse_timing}).

The default keep rate used for every dataset and backbone is listed in Sec.~C of the supplementary.

\paragraph{Keep-Rate Schedule.} Table~\ref{tab:keeprate} sweeps the per-stage keep rate $k_r$ on THUMOS-14 with the InternVideoNext-L backbone. For this model $k_r{=}0.7$ gives the best trade-off; lowering it trades accuracy for throughput gracefully, e.g., $k_r{=}0.5$ reaches 0.60 videos/s (up from 0.42) at only $-$1.13\% avg.\ mAP. \NuestroMetodo{} thus degrades smoothly rather than collapsing as more tokens are pruned, so $k_r$ can be tuned per deployment budget.

\begin{table}[t]
\centering
\caption{Effect of keep rate $k_r$ on THUMOS-14 (InternVideoNext-L backbone).}
\label{tab:keeprate}
\setlength{\tabcolsep}{4pt}
\footnotesize
\begin{tabular}{crccc}
\toprule
\textbf{Keep Rate} & \textbf{TFLOPs} & \textbf{vid/s} & \textbf{mAP@0.5} & \textbf{Avg. mAP} \\
\midrule
0.7 & 87.05 & 0.42 & 78.22 & 74.13 \\
0.6 & 74.09 & 0.50 & 77.22 & 73.64 \\
0.5 & 60.13 & 0.60 & 76.40 & 73.00 \\
\bottomrule
\end{tabular}
\end{table}

\subsection{Auxiliary Task Supervision}
\label{sec:auxiliary}

We introduce auxiliary tokens, learnable embeddings that aggregate task-specific information from visual features via cross-attention. Within each adapter, after the convolution processes the bottleneck features $\mathbf{X'_{vis}} \in \mathbb{R}^{B \times N_{kept} \times D/4}$, a lightweight cross-attention module lets auxiliary tokens query visual tokens:
\begin{equation}
    \mathbf{X}'_{aux} = \text{CrossAttn}(\mathbf{X}_{aux},\; \mathbf{X'_{vis}})
\end{equation}
where only the auxiliary tokens $\mathbf{X_{aux}}$ are updated. In this module, the tokens are further down projected ($D/8$). The queries $\mathbf{Q}_{aux}$ are projected from the auxiliary tokens $\mathbf{X_{aux}}$, while keys $\mathbf{K}_{vis}$ and values $\mathbf{V}_{vis}$ are projected from the visual features $\mathbf{X'_{vis}}$. This asymmetric design enables information flow from visual patches to auxiliary tokens without overhead on the visual token set. The cross-attention module also serves as the adapter's mechanism for modeling temporal dynamics.  

This makes our architecture flexible enough to support auxiliary task supervision without requiring per-task dense reconstructions. We demonstrate this capability with landmark-guided token selection and pose heatmap prediction, which we deploy on fine-grained assembly tasks (Sec.~\ref{sec:attach}).

\subsection{Sparse Adapter Architecture}
\label{sec:adapter}
\begin{figure}[t]
    \centering
    \includegraphics[width=0.85\linewidth]{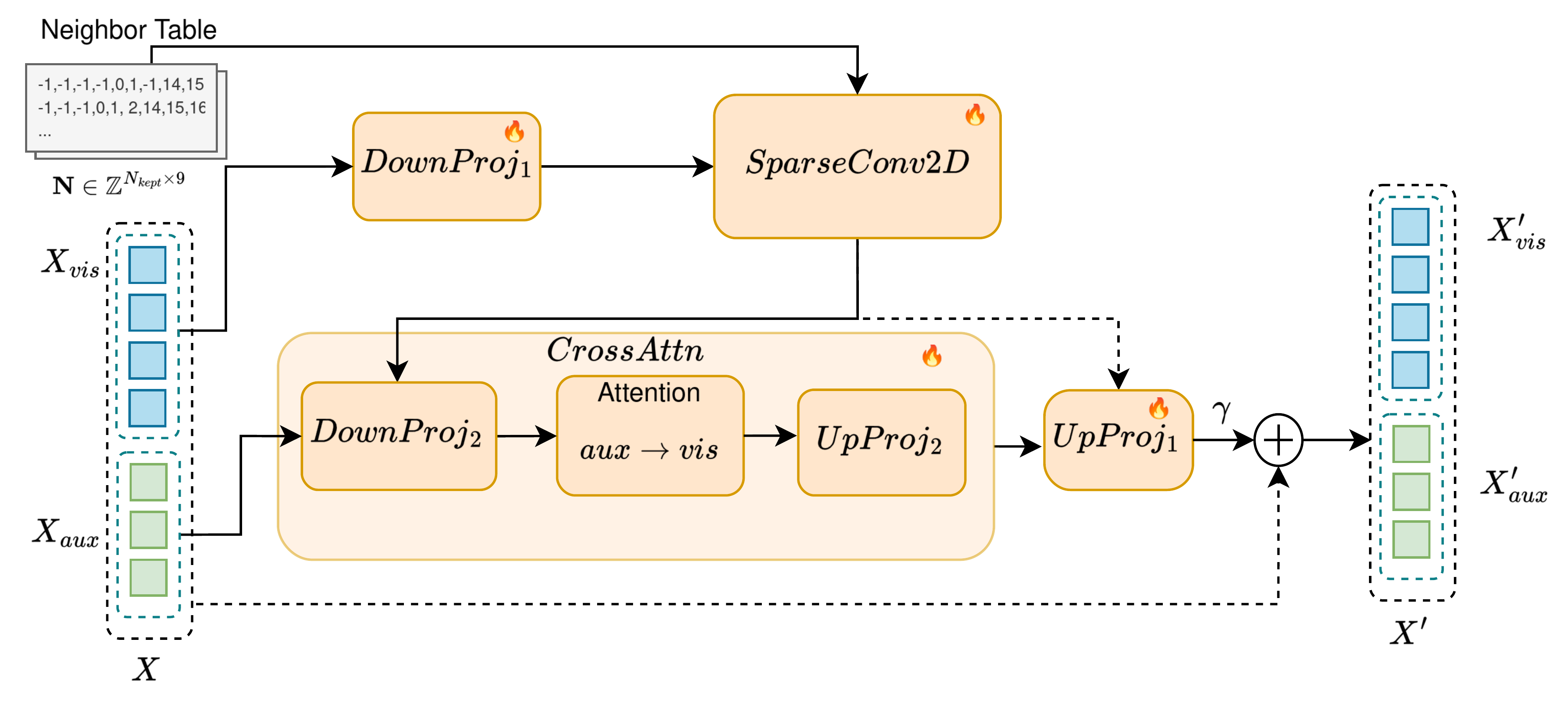}
    \caption{\textbf{Sparse adapter architecture.} Our bottleneck adapter (orange, trainable; the surrounding ViT block is frozen) is inserted after each Transformer block's attention layer. Visual ($\mathbf{X}_{vis}$) and auxiliary ($\mathbf{X}_{aux}$) tokens are down-projected ($\text{DownProj}_1$) before \sparseconv{} gathers each kept token's spatial neighbors via the precomputed neighbor index table $\mathbf{N}$ ($-1$ marks pruned/out-of-bounds neighbors, top-left inset). CrossAttn then lets auxiliary tokens query the sparse visual features, updating only $\mathbf{X}_{aux}$ at no extra cost to the visual path, before $\text{UpProj}_1$ restores the channel dimension and a learnable scalar $\gamma$ modulates the adapter's residual contribution.}

    \label{fig:adapter_diag}
\end{figure}
We integrate our sparse convolution and the auxiliary cross-attention mechanism into a bottleneck adapter, named \NuestroMetodo{}, inserted after each Transformer block's attention layer. As illustrated in Fig.~\ref{fig:adapter_diag}, input features pass through a down-projection, GELU activation, our sparse 2D convolution, and the cross-attention module, followed by an up-projection that restores the original dimension:
\begin{align}
    \mathbf{X'_{vis}} &= \text{SparseConv2D}(\sigma(\text{DownProj}_1(\mathbf{X_{vis}})),\; \mathbf{N}) \\
    \mathbf{X}' &= \text{Concat}(\text{CrossAttn}(\mathbf{X}_{aux},\; \mathbf{X'_{vis}}) , \mathbf{X'_{vis})} \\
    \mathbf{X}_{out} &= \gamma \cdot \text{UpProj}_1(\mathbf{X'}) + \mathbf{X},
\end{align}

where $DownProj_1$: $\mathbb{R}^D \rightarrow \mathbb{R}^{D/4}$, $\sigma$ is GELU, SparseConv2D uses neighbor table $\mathbf{N}$, CrossAttn lets auxiliary tokens query the sparse bottleneck features $\mathbf{X_{vis}}$, $UpProj_1$: $\mathbb{R}^{D/4} \rightarrow \mathbb{R}^D$, and $\gamma$ is a learnable scalar (initialized to 1) that controls adapter influence.
For early layers (before token selection), we use a standard 2D convolution, as our sparse convolution offers no gains in this dense regime.

\paragraph{Temporal Context Flow.}
Temporal information reaches visual tokens through two complementary pathways. First, after the adapter enriches the auxiliary tokens (e.g., CLS) at layer $\ell$, the frozen ViT self-attention at layer $\ell{+}1$ naturally propagates this enriched representation to every visual token across all frames. Second, the backbone output concatenates the final CLS token along with frame features that the detection head processes directly. The CLS thus serves both as an intermediate temporal relay between adapter layers and as a global temporal summary available to the detection head.
\section{Experiments}
\label{sec:experiments}

We demonstrate that native sparse convolutions deliver substantial efficiency gains while maintaining accuracy. We report average mAP over IoU thresholds \{0.3, 0.4, 0.5, 0.6, 0.7\} for THUMOS-14, \{0.5, 0.75, 0.95\} for ActivityNet-1.3, and \{0.1, 0.2, 0.3, 0.4, 0.5\} for ATTACH, following the standard protocol for each benchmark.

\subsection{Datasets and Setup}
\label{sec:datasets}
We evaluate \NuestroMetodo{} on 3 benchmarks for TAD: \textbf{THUMOS-14}~\cite{thumos14} (200/211 train/test videos, 20 sports classes), \textbf{ActivityNet-1.3}~\cite{caba2015activitynet} (10k/4,728 train/test videos, 200 activity classes), and \textbf{ATTACH}~\cite{attachdataset} (1,200+ assembly videos, 51 fine-grained classes with 68\% temporal overlap, person-split: 28/4/10 participants for train/val/test). For THUMOS-14 and ATTACH, we use $T=768$ frames at $224\times224$ resolution as input. For ActivityNet-1.3, $T=192$ at $160\times160$. Full dataset statistics and preprocessing details are in Sec.~B of the supplementary material.

We use frozen video transformer backbones (i.e., VideoMAEv2~\cite{wang2023videomaev2}, InternVideoNext~\cite{wang2025internvideo}), with the OpenTAD framework~\cite{liu2025opentad}. Sparse adapters are inserted after each transformer block, with token selection in blocks 4, 8, 12, 16 (ViT-L) using $k_r{=}0.6$. We employ ActionFormer~\cite{zhang2022actionformer} as the detection head.
Full hyperparameters are provided in Sec.~C of the supplementary material.

\subsection{Efficiency Analysis}
\label{sec:efficiency}
We evaluate inference of \NuestroMetodo{} (ViT-L) against AdaTAD~\cite{liuadatad2024} (ViT-L) with identical input sizes on THUMOS.
\paragraph{End-to-End Throughput.}
Figure~\ref{fig:teaser} shows an overview of TFLOPs at different model sizes.
\NuestroMetodo{} (ViT-L) processes \textbf{1.57 videos/second} vs.\ the baseline's 0.73, a \textbf{\throughputImpr{}} improvement: our native sparse convolution translates the \textbf{\TFLOPSImprLarge{}} TFLOPs reduction at $k_r{=}0.6$ directly into real-world throughput gains, validating the approach for both faster training and practical deployment.

\paragraph{Training Speed.}
Our efficiency gains also extend to training: measuring average GPU active time, \NuestroMetodo{} requires \textbf{3.01s} per step (forward$=0.70$s, backward$=2.31$s) vs.\ the baseline's 5.35s (forward$=1.49$s, backward$=3.86$s), a \trainSpeedup{} speedup. With ViT-L, where token selection prunes four times over a 24-layer backbone, \NuestroMetodo{} requires \textbf{9.23GB} peak allocated memory per training step (single RTX 3090, batch size 1) vs.\ AdaTAD's \textbf{13.30GB}, a $\sim31\%$ reduction. Despite \NuestroMetodo{} having more parameters than AdaTAD at this scale (361M vs.\ 346M), progressively shrinking the token count that reaches self-attention at deeper layers reduces activation memory enough to outweigh the parameter count, so the memory advantage of token pruning, like the TFLOPs advantage in Fig.~\ref{fig:teaser}, grows with backbone depth.

\paragraph{Kernel-Level Benchmarks.}
To verify these end-to-end gains stem from the kernel itself, we isolate the convolution module and benchmark it against dense convolution (with scatter-gather reconstruction) across temporal resolutions at $k_r{=}0.3$.

\begin{figure}[t]
    \centering
    \begin{minipage}[t]{0.48\linewidth}
        \centering
        \includegraphics[width=\linewidth]{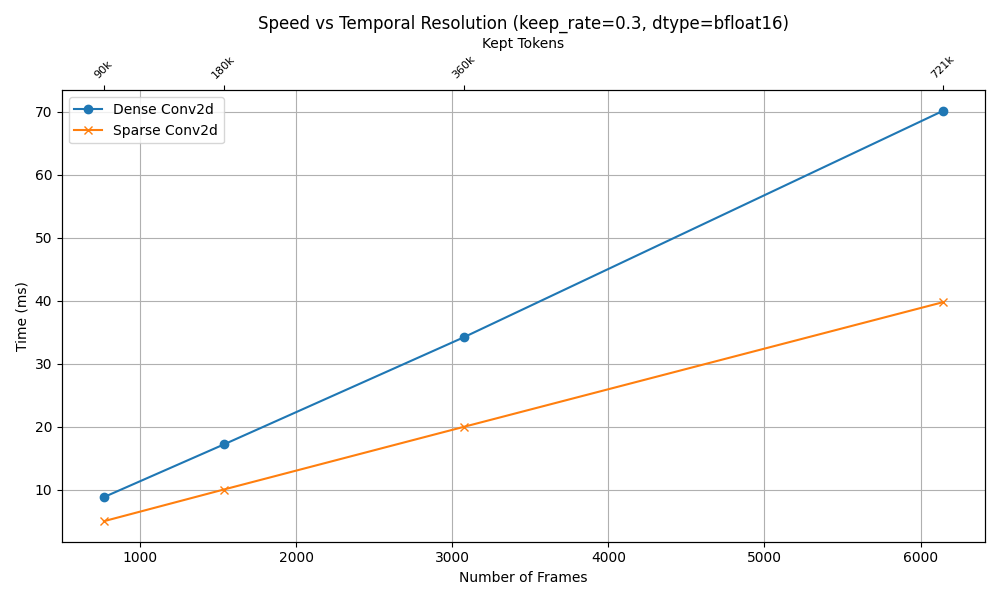}
        \subcaption{Speed scaling}
        \label{fig:speed_scaling}
    \end{minipage}
    \hfill
    \begin{minipage}[t]{0.48\linewidth}
        \centering
        \includegraphics[width=\linewidth]{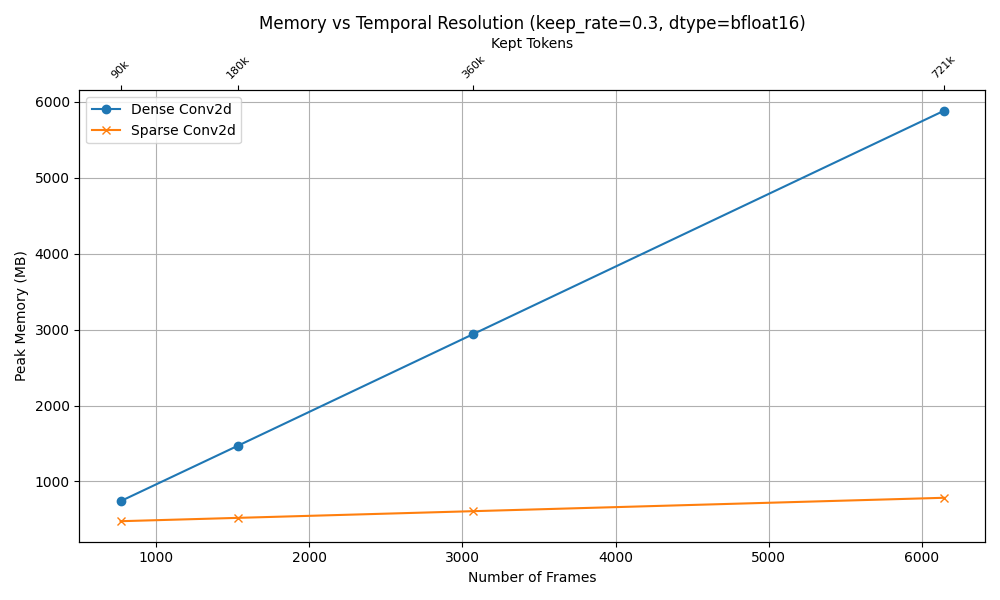}
        \subcaption{Memory scaling}
        \label{fig:memory_scaling}
    \end{minipage}
    \caption{\textbf{Sparse convolution efficiency scaling at $k_r{=}0.3$.} (a) Our sparse convolution (orange) achieves \textbf{1.75$\times$} speedup over dense (blue) at 6,144 frames (40ms vs.\ 70ms). (b) Dense convolution memory grows to $\sim$6GB at 6,144 frames regardless of sparsity. Our sparse convolution maintains \textbf{constant memory} ($\sim$800MB) via chunked processing, achieving \textbf{7.5$\times$} reduction.}
    \label{fig:efficiency_scaling}
\end{figure}

\paragraph{Kernel-Level Speed.} Fig.~\ref{fig:efficiency_scaling}(a) confirms both kernels scale linearly with sequence length, but ours operates on only $N_{kept} = k_r \cdot N$ tokens, yielding consistent speedups: \textbf{2$\times$} at 768 frames (4ms vs.\ 8ms) and \textbf{1.75$\times$} at 6,144 frames (40ms vs.\ 70ms). Since sparse adapters sit at every Transformer block, these per-layer gains compound across the network.

\paragraph{Kernel Memory Scaling.} Dense convolution materializes the full $H{\times}W{\times}C$ grid regardless of sparsity, therefore memory grows linearly with sequence length (Fig.~\ref{fig:efficiency_scaling}(b)). Ours instead allocates memory proportional to $N_{kept}$ within a fixed chunk size, holding peak memory near-constant and yielding a \textbf{87\%} reduction at 6,144 frames (800MB vs.\ 6GB), enabling sequence lengths that would otherwise run out of memory.

\subsection{Convolution Comparison: Dense vs.\ Sparse}

To isolate the effect of our sparse convolution primitive from other architectural choices (e.g.,, 1D vs.\ 2D convolution), we compare against dense convolution baselines using the same adapter architecture and token selection, with scatter-gather reconstruction. Only \sparseconv{} bypasses reconstruction entirely. For comparison, DenseConv1D is the temporal-convolution adapter identical to AdaTAD~\cite{liuadatad2024}. Table~\ref{tab:2dconv_comp} presents this comparison on THUMOS-14 with the VideoMAE-B backbone.

\begin{table}[t]
\centering
\footnotesize
\setlength{\tabcolsep}{3pt}
\begin{minipage}[t]{0.35\linewidth}
    \centering
    \caption{Dense vs.\ sparse convolution on THUMOS-14.}
    \label{tab:2dconv_comp}
    \setlength{\tabcolsep}{3pt}
    \footnotesize
    \begin{tabular}{@{}lc@{}}
    \toprule
    \textbf{Conv Type} & \textbf{Avg.} \\
    \midrule
    DenseConv1D & \first{69.60} \\
    DenseConv2D & \third{68.83} \\
    \sparseconv{} & \second{69.35} \\
    \bottomrule
    \end{tabular}
\end{minipage}\hfill
\begin{minipage}[t]{0.60\linewidth}
    \centering
    \caption{Backbone-matched comparison on THUMOS-14 with VideoMAEv2-B. All methods use identical settings. * indicates our reproduction.}
    \label{tab:backbone_matched}
    \setlength{\tabcolsep}{3pt}
    \footnotesize
    \begin{tabular}{@{}lrc@{}}
    \toprule
    \textbf{Method} & \textbf{TFLOPs} & \textbf{Avg.} \\
    \midrule
    AdaTAD* & \third{17.9} & \second{72.15} \\
    \midrule
    \NuestroMetodo{} (No CrossAttn) & \first{10.18} & \third{71.80} \\
    \NuestroMetodo{} & \second{10.29} & \first{72.44} \\
    \bottomrule
    \end{tabular}
\end{minipage}
\end{table}

\paragraph{DenseConv2D vs.\ SparseConv2D.}
Both apply a $3{\times}3$ spatial convolution over the same zero-padded grid, yet \sparseconv{} outperforms the dense baseline by \textbf{+0.52\%} average mAP. We attribute the gap to numerical divergence, where implementation differences change the  gradient accumulation trajectory over training. We verify that in FP32 without token selection ($k_r{=}1$), the gap shrinks to 0.10\%. This shows that \sparseconv{} matches 
dense accuracy while being substantially more efficient. A detailed analysis is provided in Sec.~E of the supplementary.

\paragraph{DenseConv1D vs.\ SparseConv2D.}
DenseConv1D achieves a slightly higher average mAP (\textbf{69.60} vs.\ \textbf{69.35}, $+$0.25\%), as its temporal kernel directly models inter-frame dynamics, consistent with AdaTAD~\cite{liuadatad2024}; however, this comparison is conducted under scatter-gather reconstruction, which drastically reduces the gains obtained by token selection. Implementing an efficient \emph{native sparse} 1D temporal kernel, in contrast, is difficult in practice: temporal neighbors reside in different frame blocks, causing ${\sim}$8x larger memory access distances than spatial neighbors, which breaks cache locality and coalesced warp access (see Sec.~G of the supplementary for a detailed analysis).

\subsection{Backbone-Matched Comparison}
\label{sec:backbone_matched}

To isolate the contribution of our sparse convolution framework from backbone differences, we conducted controlled experiments using the same VideoMAEv2-B backbone. Table~\ref{tab:backbone_matched} presents this fair comparison on THUMOS-14. 

\paragraph{Analysis.}
The key result is the TFLOPs column: \NuestroMetodo{} requires only \textbf{10.29 TFLOPs} (\textbf{\TFLOPSImprBase{}} fewer than AdaTAD's 17.9) while \textit{maintaining} accuracy (72.44\% vs.\ 72.15\%), confirming that native sparse convolutions eliminate reconstruction overhead without sacrificing quality and translate directly into \textbf{\throughputImpr{}} faster inference and \textbf{\trainSpeedup{}} faster training (Sec.~\ref{sec:efficiency}). 

\paragraph{Effect of Cross-Attention for Temporal Recombination.}
The variant \NuestroMetodo{} (No CrossAttn) already achieves the bulk of the computation reduction (10.18 TFLOPs) but drops slightly below AdaTAD in accuracy (71.80\% vs.\ 72.15\%). As shown in Table~\ref{tab:2dconv_comp}, replacing a temporal 1D convolution with a spatial 2D convolution sacrifices inter-frame modeling: DenseConv1D outperforms DenseConv2D by +0.77\%. Our cross-attention module compensates for this: a single CLS query aggregates information across retained patches from different frames, recovering global temporal context, and closes the gap at negligible compute cost (+0.11 TFLOPs). \sparseconv{} is also robust to the choice of token-selection criterion (e.g.,, EViT-style attention vs.\ norm-based), at matched compute (Sec.~E of the supplementary).
\subsection{Auxiliary Task Tokens Experiments}
\label{sec:attach}
\begin{table}[t]
\setlength{\tabcolsep}{3pt}
\centering
\caption{Ablation of auxiliary landmark supervision on ATTACH (Person Split) using VideoMAEv2-B.}
\begin{tabular}{lrcccccc}
\toprule
\textbf{Method} & \textbf{TFLOPs$\downarrow$} & \textbf{0.1} & \textbf{0.2} & \textbf{0.3} & \textbf{0.4} & \textbf{0.5} & \textbf{Avg.} \\
\midrule
\NuestroMetodo{} & \first{10.29} & \second{20.17} & \second{18.69} & \second{16.90} & \second{14.39} & \second{11.25} & \second{16.28} \\
\NuestroMetodo{} + Kp & \second{11.17} & \first{22.58} & \first{21.10} & \first{19.21} & \first{16.80} & \first{13.76} & \first{18.69} \\
\bottomrule
\end{tabular}
\label{sota:attach}

\end{table}
Sparse representations create a structural advantage beyond efficiency: since our framework explicitly tracks which tokens survive selection and their original grid positions, it naturally enables auxiliary task supervision. Existing convolutional adapters operate exclusively on the visual feature grid, with no mechanism to incorporate task-specific tokens. We demonstrate this on ATTACH~\cite{attachdataset}, a fine-grained assembly actions benchmark. Prior work on ATTACH evaluates only frame-level classification, we are the first to formulate it as a full TAD problem.

Table~\ref{sota:attach} shows that adding auxiliary landmark supervision (\NuestroMetodo{} + Kp, Sec.~\ref{sec:auxiliary}) improves avg.\ mAP by \textbf{+2.41\%} (18.69 vs.\ 16.28), demonstrating that the sparse adapter's cross-attention injection point is an effective channel for multitask supervision. We provide a detailed ablation of these results in the supplementary material (Sec. F). Sec.~E of the supplementary also validates that \sparseconv{} generalizes beyond TAD, to image-to-video adaptation on a frozen CLIP ViT-B/16.

\subsection{Comparison with State-of-the-Art}

\begin{table*}[t]
\centering
\caption{Results comparison on ActivityNet-1.3 and THUMOS14 (* indicates mAP obtained from official logs). Size-matched comparisons are highlighted.
}
\label{tab:combined_sota}
\footnotesize
\setlength{\tabcolsep}{3.5pt} 
\resizebox{\textwidth}{!}{
\begin{tabular}{ll r cccc r cccccc}
\toprule
& & \multicolumn{5}{c}{\textbf{ActivityNet-1.3}} & \multicolumn{7}{c}{\textbf{THUMOS14}} \\
\cmidrule(lr){3-7} \cmidrule(lr){8-14}
\textbf{Method} & \textbf{Backbone} & \textbf{TFLOPs$\downarrow$} & \textbf{0.5} & \textbf{0.75} & \textbf{0.95} & \textbf{Avg.} & \textbf{TFLOPs$\downarrow$} & \textbf{0.3} & \textbf{0.4} & \textbf{0.5} & \textbf{0.6} & \textbf{0.7} & \textbf{Avg.} \\
\midrule
\midrule
AFSD~\cite{lin2021learning} & I3D & --- & 52.40 & 35.30 & 6.50 & 34.40 & --- & 67.3 & 62.4 & 55.5 & 43.7 & 31.1 & 52.0 \\
E2E-TAD~\cite{liu2022empirical} & SlowFast-R50 & 82.48 & 50.47 & 35.99 & 10.33 & 34.10 & 82.48 & 69.4 & 64.3 & 56.0 & 46.4 & 34.9 & 54.2 \\
BasicTAD~\cite{yang2023basictad} & SlowOnly-R50 & --- & 51.20 & 33.41 & 7.57 & 33.12 & --- & 75.5 & 70.8 & 63.5 & 50.9 & 37.4 & 59.2 \\
TALLFormer~\cite{cheng2022tallformer} & VideoSwin-B & --- & 54.10 & 36.20 & 7.90 & 35.60 & --- & 76.0 & 70.0 & 63.2 & — & 34.5 & 59.4 \\
Re$^2$TAD~\cite{zhao2023retal} & Re$^2$VSwin-T & --- & 54.75 & 37.81 & 9.03 & 36.80 & --- & 77.0 & 71.5 & 62.4 & 49.7 & 36.3 & 59.4 \\
LoSA~\cite{losa2023} & VMAEv2-G & \third{132} & \second{58.5} & \third{39.8} & \third{7.80} & \third{38.6} & \second{132} & \third{85.0} & \third{81.1} & \third{74.5} & \third{65.1} & \third{49.3} & \third{71.0} \\
ViT-TAD~\cite{vit_tad} & ViT-B & --- & 55.87 & 38.47 & 8.80 & 37.40 & --- & 85.1 & 80.9 & 74.2 & 61.8 & 45.4 & 69.5 \\
AdaTAD~\cite{liuadatad2024}* & VMAE-B & \second{8.05} & \second{56.66} & \second{39.49} & \second{9.38} & \second{38.31} & \second{17.90} & \second{85.58} & \second{81.40}& \second{74.31} & \second{63.04}& \second{49.00} & \second{70.67} \\
AdaTAD~\cite{liuadatad2024}* & VMAE-L & \second{27.42} & \second{57.68} & \second{40.52} &\first{9.96} & \second{39.18} & \second{59.40} &\second{87.11}&\second{83.65} &\second{76.84}&\first{66.82} &\first{53.07} &\first{73.50} \\
AdaTAD~\cite{liuadatad2024}* & VMAEv2-G & \second{83.59} &\third{57.40} &\second{40.44}& \second{9.78}& \second{39.20}& \third{176.87} & \second{87.50}& \second{84.02}& \second{77.61}& \first{67.05}& \first{53.19}&  \second{73.87}\\
\midrule
\NuestroMetodo{} & VMAEv2-B & \first{4.75} &\first{57.09} & \first{39.97} & \first{9.77} & \first{38.80}& \first{10.29} & \first{87.54} & \first{83.24} & \first{75.28} & \first{65.54} & \first{50.61} & \first{72.44} \\
\NuestroMetodo{} & VMAEv2-L & \first{10.10} & \first{58.40} & \first{40.81} & \second{9.76} &\first{39.58} & \first{21.25} & \first{87.26} & \first{83.80} & \first{77.22}  & \second{66.21} & \second{52.85} & \second{73.47} \\
\NuestroMetodo{} & InternVideoNext-L & \first{34.00} & \first{58.79}  &\first{41.30} & \first{10.41} & \first{40.06} & \first{87.05} & \first{88.38}&\first{84.40}
& \first{78.22} & \second{66.82} & \second{52.85} & \first{74.13} \\
\bottomrule
\end{tabular}}
\end{table*}

Table~\ref{tab:combined_sota} compares \NuestroMetodo{} with recent TAD methods on ActivityNet-1.3 and THUMOS-14. The TFLOPs for \NuestroMetodo{} and AdaTAD~\cite{liuadatad2024} are for input with the same number of frames for each data set. For fair comparison, we report AdaTAD last-epoch test mAP from the authors' publicly released training logs.

In THUMOS-14, \NuestroMetodo{} with VideoMAEv2-B requires \textbf{\TFLOPSImprBase{}}  fewer TFLOPs than AdaTAD~\cite{liuadatad2024} (10.29 vs.\ 17.9) while achieving \textbf{72.44\%} average mAP, surpassing AdaTAD's 70.67\% by \textbf{+\mAPImprTHUMOS{}\%}. We note that ViT-TAD's 69.5\% with ViT-B is obtained under a substantially reduced input regime (256 frames at 160${\times}$160 vs.\ our 768 frames at 224${\times}$224), and its full fine-tuning design prevents scaling beyond ViT-B. Scaling to ViT-L widens the efficiency gap: \NuestroMetodo{} uses \textbf{\TFLOPSImprLarge{} fewer TFLOPs} (21.25 vs.\ 59.4) while matching AdaTAD's accuracy (73.47\% vs.\ 73.50\%), demonstrating that our efficiency gains grow with model size.

When scaling to InternVideoNext-L, the efficiency advantage becomes most pronounced. On THUMOS-14, \NuestroMetodo{} requires only \textbf{87.05} TFLOPs vs.\textbf{176.87}  for AdaTAD's billion-parameter VMAEv2-G, while surpassing its 73.87\% Avg. mAP by +0.26\%: a Large-sized backbone with native sparse convolutions surpasses Giant-level performance at \textbf{half the computational cost}. 
On Activity\-Net-1.3, \NuestroMetodo{} with VideoMAEv2-B uses \textbf{41\% fewer TFLOPs} (4.75 vs.\ 8.05) while achieving \textbf{38.80\%} average mAP, surpassing AdaTAD (38.39\%) by \textbf{+0.41\%}. The most compelling result is with InternVideoNext-L: \NuestroMetodo{} requires \textbf{\TFLOPSRatioInternAnet{} fewer TFLOPs} (34 vs.\ 83.59) while achieving \textbf{40.06\%} average mAP, surpassing AdaTAD with VMAEv2-G (39.20\%) by +0.86\% at every IoU threshold (Table~\ref{tab:combined_sota}). To isolate this gain from the backbone switch itself, we also run AdaTAD with the \emph{same} InternVideoNext-L backbone: it reaches 39.72\% average mAP at 95.3 TFLOPs, confirming that, at matched backbone, \NuestroMetodo{} still cuts compute substantially (34 vs.\ 95.3 TFLOPs) while improving accuracy (+0.34\%), so the gains stem from \NuestroMetodo{} itself rather than from the choice of backbone.

\section{Conclusion}

We introduced a native sparse 2D convolution primitive that enables convolutional adapters to operate on dynamically pruned token sets without reconstructing the dense spatial grid. This makes sparsity effective across the entire model rather than only within attention layers. Integrated into our SV-TAD framework, this design yields substantial reductions in compute and inference time while maintaining or improving detection accuracy across multiple benchmarks.

Crucially, the gains grow with model capacity: with ViT-L, we match top performing accuracy at \TFLOPSImprLarge{} fewer TFLOPs, while with InternVideoNext-L we set a new state of the art on both THUMOS-14 and ActivityNet-1.3, achieving the best accuracy while requiring roughly half the compute (\TFLOPSRatioInternThu{} and \TFLOPSRatioInternAnet{} fewer TFLOPs, respectively). In ATTACH, auxiliary landmark supervision enabled by the sparse adapter's cross-attention injection point, sets a new state of the art and improves our baseline model by +2.41\% mAP, demonstrating that the framework extends naturally to multitask learning on fine-grained assembly.

Beyond temporal action detection, our primitive is applicable wherever convolutional modules follow token selection in Vision Transformers, including spatial adapters, feature pyramid networks, and task-specific heads.

\paragraph{Limitations and Future Work.}
Our implementation targets NVIDIA GPUs; porting to other accelerators requires replacing the backend, although the core algorithm is architecture-agnostic. Additionally, the sparse kernel becomes faster than its dense counterpart only below the ${\sim}$55\% keep rate (Fig.~\ref{fig:sparse_timing}), so the primitive is most beneficial when progressive pruning through the layers is acceptable. Exploring learned, content-adaptive keep-rate schedules and extending the primitive to 3D spatio-temporal neighborhoods are promising directions for future work.

\section*{Acknowledgements}
This work was supported by projects PID2022-137581OB-I00, PLEC2023-010343 (INARTRANS 4.0) and PID2024-161576OB-I00, funded by the Spanish MICIU/AEI/10.13039/501100011033 and FEDER, UE, co-funded by the European Regional Development Fund (ERDF, “A way of making Europe”), and from iRoboCity2030-CM project (grant TEC-2024/TEC-62), awarded by the Community of Madrid. RP, JMB, LMB and LB are members of the Madrid ELLIS Unit, funded by the Autonomous Community of Madrid, Spain.

\bibliographystyle{plain}
\bibliography{main}

@String(CVPR  = {IEEE Conf. Comput. Vis. Pattern Recog.})

@String(ICCV  = {Int. Conf. Comput. Vis.})

@String(ECCV  = {Eur. Conf. Comput. Vis.})

@String(NeurIPS = {Adv. Neural Inform. Process. Syst.})

@String(ICLR  = {Int. Conf. Learn. Represent.})

@String(CVPR  = {CVPR})

@String(ICCV  = {ICCV})

@String(ECCV  = {ECCV})

@String(NeurIPS = {NeurIPS})

@String(ICLR  = {ICLR})

@inproceedings{yuan2024efficient,
  author       = {Xin Yuan and
                  Hongliang Fei and
                  Jinoo Baek},
  title        = {Efficient Transformer Adaptation with Soft Token Merging},
  booktitle    = {{IEEE/CVF} Conference on Computer Vision and Pattern Recognition,
                  {CVPR} 2024 - Workshops, Seattle, WA, USA, June 17-18, 2024},
  pages        = {3658--3668},
  publisher    = {{IEEE}},
  year         = {2024},
  url          = {https://doi.org/10.1109/CVPRW63382.2024.00369},
  doi          = {10.1109/CVPRW63382.2024.00369},
  bibsource    = {dblp computer science bibliography, https://dblp.org}
}

@inproceedings{kim2025faster,
  author       = {Kwonyoung Kim and
                  Jungin Park and
                  Jin Kim and
                  Hyeongjun Kwon and
                  Kwanghoon Sohn},
  title        = {Faster Parameter-Efficient Tuning with Token Redundancy Reduction},
  booktitle    = {{IEEE/CVF} Conference on Computer Vision and Pattern Recognition,
                  {CVPR} 2025, Nashville, TN, USA, June 11-15, 2025},
  pages        = {30189--30198},
  publisher    = {Computer Vision Foundation / {IEEE}},
  year         = {2025},
  url          = {https://openaccess.thecvf.com/content/CVPR2025/html/Kim\_Faster\_Parameter-Efficient\_Tuning\_with\_Token\_Redundancy\_Reduction\_CVPR\_2025\_paper.html},
  doi          = {10.1109/CVPR52734.2025.02810},
  bibsource    = {dblp computer science bibliography, https://dblp.org}
}

@article{dao2022flashattention,
  title={Flashattention: Fast and memory-efficient exact attention with io-awareness},
  author={Dao, Tri and Fu, Dan and Ermon, Stefano and Rudra, Atri and R{\'e}, Christopher},
  journal={Advances in neural information processing systems},
  volume={35},
  pages={16344--16359},
  year={2022}
}

@inproceedings{caba2015activitynet,
  author       = {Fabian Caba Heilbron and
                  Victor Escorcia and
                  Bernard Ghanem and
                  Juan Carlos Niebles},
  title        = {ActivityNet: {A} large-scale video benchmark for human activity understanding},
  booktitle    = {{IEEE} Conference on Computer Vision and Pattern Recognition, {CVPR}
                  2015, Boston, MA, USA, June 7-12, 2015},
  pages        = {961--970},
  publisher    = {{IEEE} Computer Society},
  year         = {2015},
  url          = {https://doi.org/10.1109/CVPR.2015.7298698},
  doi          = {10.1109/CVPR.2015.7298698},
  bibsource    = {dblp computer science bibliography, https://dblp.org}
}

@article{wang2025internvideo,
  title={Internvideo-next: Towards general video foundation models without video-text supervision},
  author={Wang, Chenting and Zhu, Yuhan and Xu, Yicheng and Yang, Jiange and Lin, Lang and Yan, Ziang and Wang, Yali and Wang, Yi and Wang, Limin},
  journal={arXiv preprint arXiv:2512.01342},
  year={2025}
}

@article{tay2022efficient,
  author       = {Yi Tay and
                  Mostafa Dehghani and
                  Dara Bahri and
                  Donald Metzler},
  title        = {Efficient Transformers: {A} Survey},
  journal      = {{ACM} Comput. Surv.},
  volume       = {55},
  number       = {6},
  pages        = {109:1--109:28},
  year         = {2023},
  url          = {https://doi.org/10.1145/3530811},
  doi          = {10.1145/3530811},
  bibsource    = {dblp computer science bibliography, https://dblp.org}
}

@inproceedings{beyer2022knowledge,
  author       = {Lucas Beyer and
                  Xiaohua Zhai and
                  Am{\'{e}}lie Royer and
                  Larisa Markeeva and
                  Rohan Anil and
                  Alexander Kolesnikov},
  title        = {Knowledge distillation: {A} good teacher is patient and consistent},
  booktitle    = {{IEEE/CVF} Conference on Computer Vision and Pattern Recognition,
                  {CVPR} 2022, New Orleans, LA, USA, June 18-24, 2022},
  pages        = {10915--10924},
  publisher    = {{IEEE}},
  year         = {2022},
  url          = {https://doi.org/10.1109/CVPR52688.2022.01065},
  doi          = {10.1109/CVPR52688.2022.01065},
  bibsource    = {dblp computer science bibliography, https://dblp.org}
}

@article{poguise2025,
title = {Pose-guided token selection for the recognition of activities of daily living},
journal = {Image and Vision Computing},
volume = {162},
pages = {105686},
year = {2025},
issn = {0262-8856},
doi = {https://doi.org/10.1016/j.imavis.2025.105686},
url = {https://www.sciencedirect.com/science/article/pii/S0262885625002744},
author = {Ricardo Pizarro and Roberto Valle and José M. Buenaposada and Luis M. Bergasa and Luis Baumela},
}

@inproceedings{liu2025opentad,
  author       = {Shuming Liu and
                  Chen Zhao and
                  Fatimah Zohra and
                  Mattia Soldan and
                  Alejandro Pardo and
                  Mengmeng Xu and
                  Lama Alssum and
                  Merey Ramazanova and
                  Juan Le{\'{o}}n Alc{\'{a}}zar and
                  Anthony Cioppa and
                  Silvio Giancola and
                  Carlos Hinojosa and
                  Bernard Ghanem},
  title        = {OpenTAD: {A} Unified Framework and Comprehensive Study of Temporal
                  Action Detection},
  booktitle    = {{IEEE/CVF} Conference on Computer Vision and Pattern Recognition Workshops,
                  {CVPR} Workshops 2025, Nashville, TN, USA, June 11-15, 2025},
  pages        = {2625--2635},
  publisher    = {Computer Vision Foundation / {IEEE}},
  year         = {2025},
  url          = {https://openaccess.thecvf.com/content/CVPR2025W/PVUW/html/Liu\_OpenTAD\_A\_Unified\_Framework\_and\_Comprehensive\_Study\_of\_Temporal\_Action\_CVPRW\_2025\_paper.html},
  bibsource    = {dblp computer science bibliography, https://dblp.org}
}

@inproceedings{liuadatad2024,
  author       = {Shuming Liu and
                  Chen{-}Lin Zhang and
                  Chen Zhao and
                  Bernard Ghanem},
  title        = {End-to-End Temporal Action Detection with 1B Parameters Across 1000
                  Frames},
  booktitle    = {{IEEE/CVF} Conference on Computer Vision and Pattern Recognition,
                  {CVPR} 2024, Seattle, WA, USA, June 16-22, 2024},
  pages        = {18591--18601},
  publisher    = {{IEEE}},
  year         = {2024},
  url          = {https://doi.org/10.1109/CVPR52733.2024.01759},
  doi          = {10.1109/CVPR52733.2024.01759},
  bibsource    = {dblp computer science bibliography, https://dblp.org}
}

@INPROCEEDINGS{attachdataset,
  author={Aganian, Dustin and Stephan, Benedict and Eisenbach, Markus and Stretz, Corinna and Gross, Horst-Michael},
  booktitle={2023 IEEE International Conference on Robotics and Automation (ICRA)}, 
  title={ATTACH Dataset: Annotated Two-Handed Assembly Actions for Human Action Understanding}, 
  year={2023},
  volume={},
  number={},
  pages={11367-11373},
  doi={10.1109/ICRA48891.2023.10160633}}

@article{yang2023basictad,
  author       = {Min Yang and
                  Guo Chen and
                  Yin{-}Dong Zheng and
                  Tong Lu and
                  Limin Wang},
  title        = {BasicTAD: An astounding RGB-Only baseline for temporal action detection},
  journal      = {Comput. Vis. Image Underst.},
  volume       = {232},
  pages        = {103692},
  year         = {2023},
  url          = {https://doi.org/10.1016/j.cviu.2023.103692},
  doi          = {10.1016/J.CVIU.2023.103692},
  bibsource    = {dblp computer science bibliography, https://dblp.org}
}

@inproceedings{zhao2023retal,
  author       = {Chen Zhao and
                  Shuming Liu and
                  Karttikeya Mangalam and
                  Bernard Ghanem},
  title        = {Re\({}^{\mbox{2}}\)TAL: Rewiring Pretrained Video Backbones for Reversible
                  Temporal Action Localization},
  booktitle    = {{IEEE/CVF} Conference on Computer Vision and Pattern Recognition,
                  {CVPR} 2023, Vancouver, BC, Canada, June 17-24, 2023},
  pages        = {10637--10647},
  publisher    = {{IEEE}},
  year         = {2023},
  url          = {https://doi.org/10.1109/CVPR52729.2023.01025},
  doi          = {10.1109/CVPR52729.2023.01025},
  bibsource    = {dblp computer science bibliography, https://dblp.org}
}

@inproceedings{bolya2023tome,
  author       = {Daniel Bolya and
                  Cheng{-}Yang Fu and
                  Xiaoliang Dai and
                  Peizhao Zhang and
                  Christoph Feichtenhofer and
                  Judy Hoffman},
  title        = {Token Merging: Your ViT But Faster},
  booktitle    = {The Eleventh International Conference on Learning Representations,
                  {ICLR} 2023, Kigali, Rwanda, May 1-5, 2023},
  publisher    = {OpenReview.net},
  year         = {2023},
  url          = {https://openreview.net/forum?id=JroZRaRw7Eu},
  bibsource    = {dblp computer science bibliography, https://dblp.org}
}

@inproceedings{haurum2023tokens,
  author       = {Joakim Bruslund Haurum and
                  Sergio Escalera and
                  Graham W. Taylor and
                  Thomas B. Moeslund},
  title        = {Which Tokens to Use? Investigating Token Reduction in Vision Transformers},
  booktitle    = {{IEEE/CVF} International Conference on Computer Vision, {ICCV} 2023
                  - Workshops, Paris, France, October 2-6, 2023},
  pages        = {773--783},
  publisher    = {{IEEE}},
  year         = {2023},
  url          = {https://doi.org/10.1109/ICCVW60793.2023.00085},
  doi          = {10.1109/ICCVW60793.2023.00085},
  bibsource    = {dblp computer science bibliography, https://dblp.org}
}

@inproceedings{wang2023videomaev2,
  author       = {Limin Wang and
                  Bingkun Huang and
                  Zhiyu Zhao and
                  Zhan Tong and
                  Yinan He and
                  Yi Wang and
                  Yali Wang and
                  Yu Qiao},
  title        = {VideoMAE {V2:} Scaling Video Masked Autoencoders with Dual Masking},
  booktitle    = {{IEEE/CVF} Conference on Computer Vision and Pattern Recognition,
                  {CVPR} 2023, Vancouver, BC, Canada, June 17-24, 2023},
  pages        = {14549--14560},
  publisher    = {{IEEE}},
  year         = {2023},
  url          = {https://doi.org/10.1109/CVPR52729.2023.01398},
  doi          = {10.1109/CVPR52729.2023.01398},
  bibsource    = {dblp computer science bibliography, https://dblp.org}
}

@inproceedings{zhang2022actionformer,
  author       = {Chen{-}Lin Zhang and
                  Jianxin Wu and
                  Yin Li},
  editor       = {Shai Avidan and
                  Gabriel J. Brostow and
                  Moustapha Ciss{\'{e}} and
                  Giovanni Maria Farinella and
                  Tal Hassner},
  title        = {ActionFormer: Localizing Moments of Actions with Transformers},
  booktitle    = {Computer Vision - {ECCV} 2022 - 17th European Conference, Tel Aviv,
                  Israel, October 23-27, 2022, Proceedings, Part {IV}},
  series       = {Lecture Notes in Computer Science},
  volume       = {13664},
  pages        = {492--510},
  publisher    = {Springer},
  year         = {2022},
  url          = {https://doi.org/10.1007/978-3-031-19772-7\_29},
  doi          = {10.1007/978-3-031-19772-7\_29},
  bibsource    = {dblp computer science bibliography, https://dblp.org}
}

@inproceedings{liu2022empirical,
  author       = {Xiaolong Liu and
                  Song Bai and
                  Xiang Bai},
  title        = {An Empirical Study of End-to-End Temporal Action Detection},
  booktitle    = {{IEEE/CVF} Conference on Computer Vision and Pattern Recognition,
                  {CVPR} 2022, New Orleans, LA, USA, June 18-24, 2022},
  pages        = {19978--19987},
  publisher    = {{IEEE}},
  year         = {2022},
  url          = {https://doi.org/10.1109/CVPR52688.2022.01938},
  doi          = {10.1109/CVPR52688.2022.01938},
  bibsource    = {dblp computer science bibliography, https://dblp.org}
}

@inproceedings{agrawal2025scaling,
  title={Scaling Action Detection: AdaTAD++ with Transformer-Enhanced Temporal-Spatial Adaptation},
  author={Agrawal, Tanay and Ali, Abid and Dantcheva, Antitza and Bremond, Francois},
  booktitle={Proceedings of the IEEE/CVF International Conference on Computer Vision},
  pages={12222--12231},
  year={2025}
}

@inproceedings{cheng2022tallformer,
  author       = {Feng Cheng and
                  Gedas Bertasius},
  editor       = {Shai Avidan and
                  Gabriel J. Brostow and
                  Moustapha Ciss{\'{e}} and
                  Giovanni Maria Farinella and
                  Tal Hassner},
  title        = {TallFormer: Temporal Action Localization with a Long-Memory Transformer},
  booktitle    = {Computer Vision - {ECCV} 2022 - 17th European Conference, Tel Aviv,
                  Israel, October 23-27, 2022, Proceedings, Part {XXXIV}},
  series       = {Lecture Notes in Computer Science},
  volume       = {13694},
  pages        = {503--521},
  publisher    = {Springer},
  year         = {2022},
  url          = {https://doi.org/10.1007/978-3-031-19830-4\_29},
  doi          = {10.1007/978-3-031-19830-4\_29},
  bibsource    = {dblp computer science bibliography, https://dblp.org}
}

@inproceedings{liang2022evit,
  author       = {Youwei Liang and
                  Chongjian Ge and
                  Zhan Tong and
                  Yibing Song and
                  Jue Wang and
                  Pengtao Xie},
  title        = {EViT: Expediting Vision Transformers via Token Reorganizations},
  booktitle    = {The Tenth International Conference on Learning Representations, {ICLR}
                  2022, Virtual Event, April 25-29, 2022},
  publisher    = {OpenReview.net},
  year         = {2022},
  url          = {https://openreview.net/forum?id=BjyvwnXXVn\_},
  bibsource    = {dblp computer science bibliography, https://dblp.org}
}

@inproceedings{dynamicvit,
  author       = {Yongming Rao and
                  Wenliang Zhao and
                  Benlin Liu and
                  Jiwen Lu and
                  Jie Zhou and
                  Cho{-}Jui Hsieh},
  editor       = {Marc'Aurelio Ranzato and
                  Alina Beygelzimer and
                  Yann N. Dauphin and
                  Percy Liang and
                  Jennifer Wortman Vaughan},
  title        = {DynamicViT: Efficient Vision Transformers with Dynamic Token Sparsification},
  booktitle    = {Advances in Neural Information Processing Systems 34: Annual Conference
                  on Neural Information Processing Systems 2021, NeurIPS 2021, December
                  6-14, 2021, virtual},
  pages        = {13937--13949},
  year         = {2021},
  url          = {https://proceedings.neurips.cc/paper/2021/hash/747d3443e319a22747fbb873e8b2f9f2-Abstract.html},
  bibsource    = {dblp computer science bibliography, https://dblp.org}
}

@inproceedings{tokenlearner,
  author       = {Michael S. Ryoo and
                  A. J. Piergiovanni and
                  Anurag Arnab and
                  Mostafa Dehghani and
                  Anelia Angelova},
  editor       = {Marc'Aurelio Ranzato and
                  Alina Beygelzimer and
                  Yann N. Dauphin and
                  Percy Liang and
                  Jennifer Wortman Vaughan},
  title        = {TokenLearner: Adaptive Space-Time Tokenization for Videos},
  booktitle    = {Advances in Neural Information Processing Systems 34: Annual Conference
                  on Neural Information Processing Systems 2021, NeurIPS 2021, December
                  6-14, 2021, virtual},
  pages        = {12786--12797},
  year         = {2021},
  url          = {https://proceedings.neurips.cc/paper/2021/hash/6a30e32e56fce5cf381895dfe6ca7b6f-Abstract.html},
  bibsource    = {dblp computer science bibliography, https://dblp.org}
}

@article{submanifold,
  title={Submanifold sparse convolutional networks},
  author={Graham, Benjamin and Van der Maaten, Laurens},
  journal={arXiv preprint arXiv:1706.01307},
  year={2017}
}

@inproceedings{minkowski,
  author       = {Christopher B. Choy and
                  JunYoung Gwak and
                  Silvio Savarese},
  title        = {4D Spatio-Temporal ConvNets: Minkowski Convolutional Neural Networks},
  booktitle    = {{IEEE} Conference on Computer Vision and Pattern Recognition, {CVPR}
                  2019, Long Beach, CA, USA, June 16-20, 2019},
  pages        = {3075--3084},
  publisher    = {Computer Vision Foundation / {IEEE}},
  year         = {2019},
  url          = {http://openaccess.thecvf.com/content\_CVPR\_2019/html/Choy\_4D\_Spatio-Temporal\_ConvNets\_Minkowski\_Convolutional\_Neural\_Networks\_CVPR\_2019\_paper.html},
  doi          = {10.1109/CVPR.2019.00319},
  bibsource    = {dblp computer science bibliography, https://dblp.org}
}

@article{spconv,
  author       = {Yan Yan and
                  Yuxing Mao and
                  Bo Li},
  title        = {{SECOND:} Sparsely Embedded Convolutional Detection},
  journal      = {Sensors},
  volume       = {18},
  number       = {10},
  pages        = {3337},
  year         = {2018},
  url          = {https://doi.org/10.3390/s18103337},
  doi          = {10.3390/S18103337},
  bibsource    = {dblp computer science bibliography, https://dblp.org}
}

@article{sparsetrans,
  title={Generating long sequences with sparse transformers},
  author={Child, Rewon and Gray, Scott and Radford, Alec and Sutskever, Ilya},
  journal={arXiv preprint arXiv:1904.10509},
  year={2019}
}

@inproceedings{houlsby2019adapter,
  title={Parameter-efficient transfer learning for NLP},
  author={Houlsby, Neil and Giurgiu, Andrei and Jastrzebski, Stanislaw and Morrone, Bruna and De Laroussilhe, Quentin and Gesmundo, Andrea and Attariyan, Mona and Gelly, Sylvain},
  booktitle={International conference on machine learning},
  pages={2790--2799},
  year={2019},
  organization={PMLR}
}

@inproceedings{aim,
  author       = {Taojiannan Yang and
                  Yi Zhu and
                  Yusheng Xie and
                  Aston Zhang and
                  Chen Chen and
                  Mu Li},
  title        = {{AIM:} Adapting Image Models for Efficient Video Action Recognition},
  booktitle    = {The Eleventh International Conference on Learning Representations,
                  {ICLR} 2023, Kigali, Rwanda, May 1-5, 2023},
  publisher    = {OpenReview.net},
  year         = {2023},
  url          = {https://openreview.net/forum?id=CIoSZ\_HKHS7},
  bibsource    = {dblp computer science bibliography, https://dblp.org}
}

@inproceedings{stadapter,
  author       = {Junting Pan and
                  Ziyi Lin and
                  Xiatian Zhu and
                  Jing Shao and
                  Hongsheng Li},
  editor       = {Sanmi Koyejo and
                  S. Mohamed and
                  A. Agarwal and
                  Danielle Belgrave and
                  K. Cho and
                  A. Oh},
  title        = {ST-Adapter: Parameter-Efficient Image-to-Video Transfer Learning},
  booktitle    = {Advances in Neural Information Processing Systems 35: Annual Conference
                  on Neural Information Processing Systems 2022, NeurIPS 2022, New Orleans,
                  LA, USA, November 28 - December 9, 2022},
  year         = {2022},
  url          = {http://papers.nips.cc/paper\_files/paper/2022/hash/a92e9165b22d4456fc6d87236e04c266-Abstract-Conference.html},
  bibsource    = {dblp computer science bibliography, https://dblp.org}
}

@inproceedings{stts,
  author       = {Junke Wang and
                  Xitong Yang and
                  Hengduo Li and
                  Li Liu and
                  Zuxuan Wu and
                  Yu{-}Gang Jiang},
  editor       = {Shai Avidan and
                  Gabriel J. Brostow and
                  Moustapha Ciss{\'{e}} and
                  Giovanni Maria Farinella and
                  Tal Hassner},
  title        = {Efficient Video Transformers with Spatial-Temporal Token Selection},
  booktitle    = {Computer Vision - {ECCV} 2022 - 17th European Conference, Tel Aviv,
                  Israel, October 23-27, 2022, Proceedings, Part {XXXV}},
  series       = {Lecture Notes in Computer Science},
  volume       = {13695},
  pages        = {69--86},
  publisher    = {Springer},
  year         = {2022},
  url          = {https://doi.org/10.1007/978-3-031-19833-5\_5},
  doi          = {10.1007/978-3-031-19833-5\_5},
  bibsource    = {dblp computer science bibliography, https://dblp.org}
}

@inproceedings{vit_tad,
  author       = {Min Yang and
                  Huan Gao and
                  Ping Guo and
                  Limin Wang},
  title        = {Adapting Short-Term Transformers for Action Detection in Untrimmed
                  Videos},
  booktitle    = {{IEEE/CVF} Conference on Computer Vision and Pattern Recognition,
                  {CVPR} 2024, Seattle, WA, USA, June 16-22, 2024},
  pages        = {18570--18579},
  publisher    = {{IEEE}},
  year         = {2024},
  url          = {https://doi.org/10.1109/CVPR52733.2024.01757},
  doi          = {10.1109/CVPR52733.2024.01757},
  bibsource    = {dblp computer science bibliography, https://dblp.org}
}

@inproceedings{losa2023,
  author       = {Akshita Gupta and
                  Gaurav Mittal and
                  Ahmed Magooda and
                  Ye Yu and
                  Graham W. Taylor and
                  Mei Chen},
  title        = {LoSA: Long-Short-Range Adapter for Scaling End-to-End Temporal Action
                  Localization},
  booktitle    = {{IEEE/CVF} Winter Conference on Applications of Computer Vision, {WACV}
                  2025, Tucson, AZ, USA, February 26 - March 6, 2025},
  pages        = {2092--2102},
  publisher    = {{IEEE}},
  year         = {2025},
  url          = {https://doi.org/10.1109/WACV61041.2025.00210},
  doi          = {10.1109/WACV61041.2025.00210},
  bibsource    = {dblp computer science bibliography, https://dblp.org}
}

@inproceedings{lin2021learning,
  author       = {Chuming Lin and
                  Chengming Xu and
                  Donghao Luo and
                  Yabiao Wang and
                  Ying Tai and
                  Chengjie Wang and
                  Jilin Li and
                  Feiyue Huang and
                  Yanwei Fu},
  title        = {Learning Salient Boundary Feature for Anchor-free Temporal Action
                  Localization},
  booktitle    = {{IEEE} Conference on Computer Vision and Pattern Recognition, {CVPR}
                  2021, virtual, June 19-25, 2021},
  pages        = {3320--3329},
  publisher    = {Computer Vision Foundation / {IEEE}},
  year         = {2021},
  doi          = {10.1109/CVPR46437.2021.00333},
  bibsource    = {dblp computer science bibliography, https://dblp.org}
}

@inproceedings{cubuk2020randaugment,
  title={Randaugment: Practical automated data augmentation with a reduced search space},
  author={Cubuk, Ekin D and Zoph, Barret and Shlens, Jonathon and Le, Quoc V},
  booktitle={Proceedings of the IEEE/CVF conference on computer vision and pattern recognition workshops},
  pages={702--703},
  year={2020}
}

@inproceedings{lin2019bmn,
  author       = {Tianwei Lin and
                  Xiao Liu and
                  Xin Li and
                  Errui Ding and
                  Shilei Wen},
  title        = {{BMN:} Boundary-Matching Network for Temporal Action Proposal Generation},
  booktitle    = {2019 {IEEE/CVF} International Conference on Computer Vision, {ICCV}
                  2019, Seoul, Korea (South), October 27 - November 2, 2019},
  pages        = {3888--3897},
  publisher    = {{IEEE}},
  year         = {2019},
  url          = {https://doi.org/10.1109/ICCV.2019.00399},
  doi          = {10.1109/ICCV.2019.00399},
  bibsource    = {dblp computer science bibliography, https://dblp.org}
}

@article{thumos14,
  author       = {Haroon Idrees and
                  Amir R. Zamir and
                  Yu{-}Gang Jiang and
                  Alex Gorban and
                  Ivan Laptev and
                  Rahul Sukthankar and
                  Mubarak Shah},
  title        = {The {THUMOS} challenge on action recognition for videos "in the wild"},
  journal      = {Comput. Vis. Image Underst.},
  volume       = {155},
  pages        = {1--23},
  year         = {2017},
  url          = {https://doi.org/10.1016/j.cviu.2016.10.018},
  doi          = {10.1016/J.CVIU.2016.10.018},
  bibsource    = {dblp computer science bibliography, https://dblp.org}
}

@inproceedings{li2018resound,
  title={Resound: Towards action recognition without representation bias},
  author={Li, Yingwei and Li, Yi and Vasconcelos, Nuno},
  booktitle={Proceedings of the European Conference on Computer Vision (ECCV)},
  pages={513--528},
  year={2018}
}

\clearpage
\begin{center}
\textbf{\large Supplemental Materials}
\end{center}
\setcounter{section}{0}
\renewcommand{\thesection}{S-\Roman{section}}
\setcounter{equation}{0}
\setcounter{figure}{0}
\setcounter{table}{0}
\setcounter{page}{1}
\makeatletter
\renewcommand{\thetable}{S\arabic{table}}
\renewcommand{\theequation}{S\arabic{equation}}
\renewcommand{\thefigure}{S\arabic{figure}}

This supplementary provides: CUDA kernel implementation details (Sec.~\ref{sec:supp_cuda}), dataset descriptions (Sec.~\ref{sec:supp_datasets}), full training hyperparameters (Sec.~\ref{sec:supp_training}), AdaTAD++ reproduction analysis (Sec.~\ref{sec:supp_adatad}), additional ablations including keep rate sensitivity, dense vs.\ sparse numerical divergence, robustness to the selection criterion, and generality beyond TAD (Sec.~\ref{sec:supp_ablations}), detailed ATTACH analysis (Sec.~\ref{sec:supp_attach}), and a discussion covering the incompatibility between token selection and standard convolutions, dense vs.\ sparse convolution comparisons, and why a native sparse 1D temporal kernel is infeasible..

\section{CUDA Kernel Implementation Details}
\label{sec:supp_cuda}

We implement two complementary kernel strategies, automatically selected based on the channel dimension $C_{out}$. Both share the same interface: they consume the neighbor index table $\mathbf{N}$ (Sec.~3.2 of the main paper) and produce outputs only at kept positions, with no dense reconstruction.

\subsection{Implicit GEMM via cuBLAS ($C \geq 256$)}
A direct implementation would iterate over all nine $3{\times}3$ kernel positions, issuing a separate GEMM, gather, and scatter per position, 36 kernel launches with substantial host-side overhead. We instead reorganise the data layout so that a single strided batched GEMM replaces all nine multiplications. The forward pass reduces to three kernel launches:
\begin{enumerate}
    \item \textbf{Fused gather with implicit zero-padding}: A single CUDA kernel populates $\mathbf{X}_g \in \mathbb{R}^{9 \times N \times C_{in}}$ by gathering all 9 neighbors for all $N$ tokens simultaneously. Invalid neighbors ($\mathbf{N}[k,j] = -1$) are written as zeros directly.
    \item \textbf{Single batched GEMM}: The layout $[9, N, C_{in}]$ is specifically designed for \texttt{cublasGemmStridedBatchedEx}, which computes all 9 matrix multiplications $\mathbf{Y}_j = \mathbf{X}_g[j] \cdot \mathbf{W}_j$ in one call, fully utilizing Tensor Cores.
    \item \textbf{Fused reduction}: A parallel reduction sums over the nine kernel positions, adds bias, and casts to the output dtype. This replaces scatter-based accumulation (\texttt{index\_add\_}) with a simple strided read, eliminating all CPU synchronisation points.
\end{enumerate}
Chunked processing extends naturally: tokens are partitioned into fixed-size chunks (e.g.,, 32K), bounding peak memory independently of sequence length.
\subsection{Custom Tiled Kernel ($C < 256$)}
When $C_{out}$ is small, cuBLAS launch overhead dominates compute. We therefore provide hand-tuned CUDA kernels that fuse gather and multiply--accumulate into a single launch with no intermediate buffers, following a tiled GEMM structure adapted for sparse convolution:
\begin{enumerate}
    \item \textbf{Tile configuration}: Each thread block processes a tile of $32$ output tokens $\times$ $32$ output channels. Threads are mapped so each computes $2$ tokens $\times$ $4$ channels (a \texttt{float4}), achieving 2$\times$ thread coarsening that amortizes neighbor index lookups and register pressure.

    \item \textbf{Cooperative loading}: In the load phase, threads collaboratively fetch a $32 \times 32$ tile of input features and the corresponding weight slice into shared memory. The sparse gather happens here: each thread reads its assigned row's neighbor index from the indirection table and fetches from the neighbor's location (or writes zero if $-1$). This fuses gather with the tile load at no extra cost.

    \item \textbf{Compute phase}: A fully-unrolled inner loop multiplies the shared memory tiles, accumulating into \texttt{float4} registers. The outer loops iterate over the 9 kernel positions and $C_{in}/32$ input channel tiles.
\end{enumerate}

\paragraph{Sensitivity to spatial distribution.}
While the sparse kernels outperform dense baselines even under uniform random pruning, they reach peak throughput when retained tokens are spatially clustered. In practice, our token selection modules learn to concentrate tokens around salient actors and objects, so threads within a warp access neighbouring memory addresses, maximising cache-line reuse and memory coalescing.
\subsection{Shared Kernel Infrastructure}

\paragraph{Vectorised memory access.}
All kernels use 128-bit loads (\texttt{float4} for FP32, \texttt{int4} for FP16/BF16) with specialised unpack routines per dtype to maximise memory bandwidth.

\paragraph{Mixed-precision accumulation.}
Inputs in FP16/BF16 are accumulated in FP32 registers. A fused operation adds bias, casts to the output dtype, and writes results to global memory in a single pass.
\subsection{Complexity Analysis}
\label{sec:supp_complexity}

Our sparse convolution requires $\mathcal{O}(N_{kept} \cdot K^2 \cdot C_{in} \cdot C_{out})$ operations versus $\mathcal{O}(THW \cdot K^2 \cdot C_{in} \cdot C_{out})$ for the dense baseline, yielding a direct $k_r{\times}$ compute reduction. Chunked processing further decouples peak memory from sequence length: with a fixed chunk of 32K tokens, memory is bounded to $\mathcal{O}(\text{chunk\_size} \cdot (C_{in} + C_{out}))$, enabling arbitrarily long video sequences. Since the full model applies token selection at $S$ stages (e.g.,, $S{=}3$ at blocks 4, 7, 10 for ViT-B), layers after the final selection stage operate at the most aggressive sparsity, where per-layer gains are largest.

\section{Dataset Details}
\label{sec:supp_datasets}

\subsection{THUMOS-14}
THUMOS-14~\cite{thumos14} contains 200 validation and 211 test videos from 20 sports action classes, totaling approximately 26 hours of video at 30 fps ($\sim$2.8M frames). Videos are untrimmed with an average of 15 action instances per video, exhibiting significant temporal overlap and diverse action durations ranging from fractions of a second to minutes. Following AdaTAD~\cite{liuadatad2024}, we sample 768 frames per video with temporal stride 1, chunking each into 48 non-overlapping 16-frame clips processed by the ViT backbone. We follow standard protocol: train on validation set, evaluate on test set.

\subsection{ActivityNet-1.3}
ActivityNet-1.3~\cite{caba2015activitynet} contains 10,024 training, 4,926 validation, and 5,044 test videos spanning 200 activity classes, comprising approximately 849 hours ($\sim$91M frames at 30 fps). As test labels are not publicly released, we train on the training split and evaluate on the validation split, excluding 198 videos blocked on YouTube following the convention of BMN~\cite{lin2019bmn}, yielding 4,728 usable videos for evaluation. Videos average 2 minutes with 1.5 action instances each. Due to longer durations, we sample 192 frames uniformly across each video with adaptive stride, yielding 12 clips regardless of original length.

\subsection{ATTACH}
ATTACH~\cite{attachdataset} features fine-grained bimanual assembly tasks at 30 fps, totaling $\sim$40 hours ($\sim$4.3M frames). The dataset contains 51 action classes across 1,200+ videos with dense temporal annotations. We sample 768 frames per video with stride 4, covering $\sim$100 seconds per video. Over 68\% of annotations temporally overlap due to simultaneous two-handed activities.

\section{Full Training Configuration}
Table~\ref{tab:supp_hyperparams} summarizes the complete hyperparameters used across all three datasets.
\label{sec:supp_training}
\begin{table}[t]
\setlength{\tabcolsep}{1pt}
\caption{Complete training hyperparameters across all datasets.}

\footnotesize
\centering
\label{tab:supp_hyperparams}
\begin{tabular}{lccc}
\toprule
\textbf{Hyperparameter} & \textbf{THUMOS-14} & \textbf{ActivityNet} & \textbf{ATTACH} \\
\midrule
Epochs & 100 & 10 & 50 \\
Batch size  & 2 & 16 & 4 \\
Number of GPUs & 2 & 4 & 2 \\
Optimizer & \multicolumn{3}{c}{AdamW} \\
Base learning rate & \multicolumn{3}{c}{$1 \times 10^{-4}$} \\
Weight decay & \multicolumn{3}{c}{$5 \times 10^{-2}$} \\
LR schedule & \multicolumn{3}{c}{Cosine} \\
Warmup epochs & \multicolumn{3}{c}{5} \\
Precision & \multicolumn{3}{c}{BFloat16} \\
\midrule
Frames sampled & 768 & 192 & 768 \\
Temporal stride & 1 & adaptive & 4 \\
Spatial resolution & $224 \times 224$ & $160 \times 160$ & $224 \times 224$ \\
\midrule
Adapter bottleneck ratio & \multicolumn{3}{c}{4} \\
Sparse conv kernel & \multicolumn{3}{c}{$3 \times 3$} \\
Token sel. blocks (ViT-B) & \multicolumn{3}{c}{4, 7, 10} \\
Token sel. blocks (ViT-L) & \multicolumn{3}{c}{4, 8, 12, 16} \\
Default keep rate $k_r$ & \multicolumn{3}{c}{0.6 (VMAE), 0.7 (InternVideo)} \\
\midrule
Auxiliary loss weight $\lambda$ & - & - & 2.0 \\
Number of keypoints & - & - & 17 \\
Heatmap resolution & - & - & $56 \times 56$ \\
\bottomrule
\end{tabular}
\end{table}

\paragraph{Data Augmentation.}
We apply random temporal cropping (crop ratio 0.9--1.0), random horizontal flipping (probability 0.5), and RandAugment~\cite{cubuk2020randaugment}.

\paragraph{Detection Head.}
ActionFormer with 6 pyramid levels, regression range ratios $\{1,2,4,8,16,32\}$, focal loss ($\gamma{=}2$, $\alpha{=}0.25$) for classification, and DIoU loss for regression.

\subsection{End-to-End Training and Inference}
\label{sec:e2e}

Our complete pipeline integrates a frozen pretrained\cite{wang2023videomaev2,wang2025internvideo} backbone with trainable sparse adapters and an ActionFormer~\cite{zhang2022actionformer} detection head. During training, we optimize only the sparse adapters and detection head, keeping the backbone frozen for parameter efficiency. When auxiliary supervision is used (e.g.,, on ATTACH), the auxiliary head is jointly optimized with loss $\mathcal{L} = \mathcal{L}_{det} + \lambda \mathcal{L}_{aux}$, where $\mathcal{L}_{aux}$ is MSE for heatmap regression. Otherwise, training uses $\mathcal{L} = \mathcal{L}_{det}$ alone.
\section{AdaTAD++ Reproduction Analysis}
\label{sec:supp_adatad}

We attempted to reproduce AdaTAD++~\cite{agrawal2025scaling} results using the OpenTAD framework~\cite{liu2025opentad}. Table~\ref{tab:supp_reproduction} summarizes our reproduction attempt.

\begin{table}[t]
\centering
\caption{AdaTAD++ reproduction comparison. Reported values from original paper vs.\ our reproduction using OpenTAD.}
\label{tab:supp_reproduction}
\begin{tabular}{llcccccc}
\toprule
& & \multicolumn{6}{c}{\textbf{THUMOS-14 (mAP@IoU)}} \\
\cmidrule(lr){3-8}
\textbf{Method} & \textbf{Source} & \textbf{0.3} & \textbf{0.4} & \textbf{0.5} & \textbf{0.6} & \textbf{0.7} & \textbf{Avg.} \\
\midrule
AdaTAD++ (VMAE-B) & Reported & 88.3 & 83.7 & 76.6 & 65.0 & 51.8 & 73.3 \\
AdaTAD++ (VMAE-B) & Ours & 84.64 & 80.33 & 72.68 & 62.44 & 49.67 & 69.95 \\
\bottomrule
\end{tabular}
\end{table}

\paragraph{Reproduction Gap Analysis.}
Our AdaTAD++ reproduction achieves 69.95\% vs.\ 73.3\% reported ($-$3.35\%). We attribute this discrepancy to:
\begin{itemize}
    \item \textbf{Missing hyperparameters}: The original AdaTAD++ paper does not release code or full hyperparameter configurations.
    \item \textbf{Data augmentation}: Specific augmentation strategies are not fully documented.
    \item \textbf{Training dynamics}: Training techniques are not fully specified.
\end{itemize}

Despite extensive hyperparameter search, we were unable to close this gap. Our reproduction used the publicly available AdaTAD~\cite{liuadatad2024} codebase as the base configuration, applying the additional AdaTAD++ modifications as described in the paper. We contacted the authors to request the missing configurations but received no reply.

\section{Additional Ablations}
\label{sec:supp_ablations}

\subsection{Dense vs.\ Sparse 2D Convolution: Numerical Divergence}
\label{sec:supp_dense_vs_sparse}

Although a dense 2D convolution over a zero-padded grid and our sparse 2D convolution are mathematically equivalent, \sparseconv{} outperforms the dense baseline by +0.52\% average mAP on THUMOS-14 (Table~2 of the main paper). We identify two sources of numerical divergence that explain this gap.

\paragraph{Different forward computational backends.}
Our sparse convolution dispatches a cuBLAS strided batched GEMM with explicit FP32 accumulation (\texttt{cuda\_\-compute\_type = CUDA\_R\_32F}), whereas the dense path uses cuDNN's \texttt{conv2d}, which may employ different tiling strategies, Winograd transforms, or accumulation orders internally. Because floating-point addition is non-associative, these differing reduction orders produce slightly different rounding trajectories even when the mathematical operation is identical. Under BFloat16 inputs, per-element quantization amplifies these trajectory differences with every forward pass.

\paragraph{Sparse backward kernel skips pruned positions.}
The backward kernels of our sparse convolution (both \texttt{sparse\_bwd\_in} and \texttt{sparse\_bwd\_w}) explicitly skip pruned positions via the $-1$ sentinel in the neighbor table (\texttt{if (n\_idx == -1) continue}). The dense backward, dispatched through PyTorch autograd on \texttt{F.\-conv2d}, accumulates weight-gradient contributions from all $H{\times}W$ grid positions including the zero-padded ones. Although these zero-input contributions are analytically zero, excluding them changes the number of terms in the accumulation and their ordering, yielding marginally different weight updates that compound over thousands of training steps.

\paragraph{Verification.}
We validate this explanation with two controlled experiments. First, we compare kernel-level outputs at different precision levels: at FP32, dense and sparse outputs match to within machine epsilon, whereas under B\-Float16 discrepancies emerge and increase as the keep rate decreases, consistent with greater backend divergence when more positions differ between the two paths. Second, we eliminate both sources simultaneously by running both methods at $k_r{=}1.0$ (no pruning) in FP32, making the computational graphs equivalent. Under these conditions, \sparseconv{} obtains 69.79\% and Dense 2DConv obtains 69.89\%, a 0.10\% gap well within run-to-run variance. This confirms that the +0.52\% gap observed at $k_r{=}0.6$ under BFloat16 is a reproducible consequence of the backend and gradient-path differences described above, not a methodological artifact.

\subsection{Robustness to the Token Selection Criterion}
\label{sec:supp_selector}
\sparseconv{} consumes only the kept-token grid coordinates, so it is agnostic to the criterion used to choose which tokens are kept. To test this, we swap only the selector on THUMOS-14 with VideoMAEv2-B, keeping every other component fixed: an EViT-style attention criterion~\cite{liang2022evit} reaches 72.43\%/10.301 TFLOPs and a norm-based criterion (ranking tokens by post-LayerNorm activation magnitude, following the survey of selection criteria in~\cite{haurum2023tokens}) reaches 68.06\%/10.296 TFLOPs, against our default attention-based selector at 72.44\%/10.297 TFLOPs. The near-identical compute confirms that \sparseconv{} adds no overhead regardless of selector; the residual accuracy gap is determined by the selection criterion itself, not by the sparse convolution.

\subsection{Generality Beyond Temporal Action Detection}
\label{sec:supp_generality}
\sparseconv{} is designed as a general token-selection primitive for PEFT adap\-ters in ViTs, not one specific to TAD or to the \NuestroMetodo{} architecture. To test cross-task and cross-backbone applicability, we integrate it into image-to-video adaptation: a \emph{frozen CLIP ViT-B/16} fine-tuned with AIM~\cite{aim} for action recognition on Diving-48~\cite{li2018resound}. We replace AIM's S\_Adapter, a dense spatial adapter, with \sparseconv{} combined with token selection, leaving the rest of the AIM recipe unchanged. Our variant reaches \textbf{88.7\%} top-1 accuracy at \textbf{1.01} TFLOPs, against AIM's \textbf{88.9\%} at \textbf{1.62} TFLOPs, a \textbf{37.5\%} compute reduction at comparable accuracy ($-$0.2\%). This shows that the efficiency benefit of \sparseconv{} is not specific to temporal action detection: it transfers directly to a different task (action classification) and a different frozen backbone (CLIP ViT-B/16).

\section{Detailed ATTACH Analysis}
\label{sec:supp_attach}

ATTACH poses a distinct challenge for sparse methods: over 68\% of its annotations temporally overlap due to simultaneous bimanual actions, requiring the model to represent multiple concurrent activities per frame. Table~\ref{sota:attach_full} ablates our design choices on this benchmark.

\begin{table}
\setlength{\tabcolsep}{3pt}
\centering
\caption{Ablation on ATTACH (Person Split) with VideoMAEv2-B. \NuestroMetodo{} + Kp adds auxiliary landmark supervision. * denotes our reproduction.}
\begin{tabular}{lcccccc}
\toprule
\textbf{Method} & \textbf{0.1} & \textbf{0.2} & \textbf{0.3} & \textbf{0.4} & \textbf{0.5} & \textbf{Avg.} \\
\midrule
AdaTAD*~\cite{liuadatad2024} & 21.68 & 20.20 & 18.32  & 16.00 & 13.03& 17.84\\
\midrule
\NuestroMetodo{} (No CrossAttn)&21.11 & 19.53 & 17.52& 15.09 & 12.20 & 17.09 \\
\NuestroMetodo{}  & 20.17 & 18.69 &16.90 & 14.39 & 11.25 & 16.28 \\
\NuestroMetodo{} + Kp & \textbf{22.58} & \textbf{21.10} & \textbf{19.21} & \textbf{16.80} & \textbf{13.76} & \textbf{18.69} \\
\bottomrule
\end{tabular}
\label{sota:attach_full}
\end{table}

\paragraph{Effect of cross-attention without auxiliary tokens.}
Comparing rows~2 and~3 isolates the cross-attention module: adding it \emph{reduces} performance from 17.09\% to 16.28\% ($-$0.81\%). Under sparsity, a single CLS query must aggregate information for multiple concurrent actions, creating a representational bottleneck that is particularly harmful on multilabel data where frames routinely contain two simultaneous hand actions.

\paragraph{Landmark supervision resolves this bottleneck.}
Replacing the single CLS query with $K{=}17$ pose-landmark queries provides spatially-grounded attention targets, each tied to a specific body region. \NuestroMetodo{} + Kp achieves \textbf{18.69\%} avg.\ mAP, surpassing AdaTAD by \textbf{+0.85\%} while requiring 38\% fewer FLOPs (11.17 vs.\ 17.90 TFLOPs). Assembly actions are defined by hand and arm movements, so pose landmarks directly inform the disambiguation of dense, overlapping action boundaries. As a secondary benefit, the auxiliary loss guides token selection toward task-relevant regions (i.e., hands and manipulated objects), improving both feature quality and computational focus.

\paragraph{Summary.}
These results show that the cross-attention module is beneficial only when paired with informative auxiliary tokens. On datasets lacking such supervision, the simpler \textit{No CrossAttn} variant is preferable. More broadly, they demonstrate the modularity of our framework: domain-specific auxiliary tasks can be incorporated with minimal architectural overhead when suitable supervision is available, and cleanly removed otherwise.

\section{Discussion}
\label{sec:supp_discussion}
We address three design questions that arise from the main paper: whether memory overhead increases (Sec.~\ref{sec:supp_memory}), why standard convolutions cannot directly operate on sparse tokens (Sec.~\ref{sec:supp_scatter_gather}), and why we chose a spatial rather than temporal sparse kernel (Sec.~\ref{sec:supp_memory_locality})..
\subsection{Memory Consumption}
\label{sec:supp_memory}

Peak GPU memory remains comparable between \NuestroMetodo{} and AdaTAD. The frozen backbone activations dominate the overall memory footprint regardless of adapter design. At 768 frames with VideoMAEv2-Base, the backbone accounts for the vast majority of allocated memory, while the adapter modules (whether dense or sparse) contribute a negligible fraction. Because our sparse kernels replace rather than augment the convolution pathway, they do not introduce additional buffers beyond the neighbor index table ($N_{kept} \times 9$ int32 entries, $\sim$0.2~MB at $k_r{=}0.6$). In practice, \NuestroMetodo{} and AdaTAD occupy nearly the same peak memory, meaning the throughput and FLOP reductions come at no memory cost.

\subsection{Incompatibility Between Token Selection and Standard Convolutions}
\label{sec:supp_scatter_gather}
Standard convolutions assume a \textit{dense, regularly structured} input tensor. Token selection breaks this assumption  by producing a variable-length list of kept tokens rather than a dense grid. After global TopK pruning~\cite{liang2022evit} across the flattened spatio-temporal sequence $(B, T{\times}H{\times}W, C)$, the output is a variable-length list of kept tokens, not a dense grid. This creates two structural problems: (i)~\textbf{non-uniform token counts per frame} (frames with salient content retain more tokens, violating the fixed $H{\times}W$ layout that \texttt{Conv1d}/\texttt{Conv2d} expect), and (ii)~\textbf{misaligned spatial positions across frames} (the \textit{which} positions survive differs per frame, so grid indices in the sparse list have no consistent spatial correspondence).

\paragraph{The scatter-gather workaround.}
A solution is to reconstruct the dense tensor before convolution: \textit{scatter} the $N_{kept}$ sparse tokens back to their original grid positions in a zero-initialized tensor of full size, apply the convolution, then \textit{gather} outputs at the kept positions. Concretely, for a 1D temporal adapter, this involves:

\begin{enumerate}
    \item \textbf{Allocate} a dense tensor $\mathbf{X}_{dense} \in \mathbb{R}^{B \times H \times W \times T \times C}$, initialized to zero.
    \item \textbf{Scatter}: for each kept token $k$ at grid position $(b, t, h, w)$, write \\ $\mathbf{X}_{dense}[b, h, w, t, :] = \mathbf{x}_k$.
    \item \textbf{Reshape} to $(B{\times}H{\times}W, C, T)$ and apply \texttt{nn.Conv1d}.
    \item \textbf{Gather}: extract outputs at the $N_{kept}$ positions via \texttt{gather}, discarding the zero-padded results.
\end{enumerate}

For a 2D spatial adapter, the process is analogous: \\ scatter into $(B{\times}T, C, H, W)$, apply \texttt{nn.Conv2d}, and gather back. The key point is that \textit{every} standard convolution variant requires this same reconstruct-convolve-extract pipeline.

\paragraph{Efficiency limitations.}
This workaround has three compounding costs. \textbf{(1)~Memory}: the dense tensor allocates $\mathcal{O}(H{\times}W{\times}T{\times}C)$ regardless of how many tokens were pruned, growing linearly with frame count. \textbf{(2)~Wasted computation}: the convolution operates over the full tensor including zero-padded positions; at $k_r{=}0.6$, $\sim$40\% of FLOPs produce no useful output. \textbf{(3)~Scatter-gather overhead}: both operations involve irregular memory access patterns that create a hard latency floor, pruning more aggressively reduces convolution FLOPs but cannot reduce this fixed cost.
\paragraph{Native sparse convolution.}
\sparseconv{} replaces the entire allocate-scatter-convolve-gather pipeline with a single fused operation. The precomputed neighbor index table $\mathbf{N} \in \mathbb{Z}^{N_{kept} \times 9}$ maps each kept token directly to its neighbors in the sparse list. The CUDA kernel gathers features through this indirection, computes the convolution, and writes outputs, all without materializing a dense grid. Memory and computation scale with $N_{kept}$, not $H{\times}W{\times}T$, converting token reduction ratios into proportional wall-clock savings.

\subsection{Why Spatial Rather Than Temporal Sparse Convolution}
\label{sec:supp_memory_locality}

Dense 1DConv achieves marginally higher mAP than Dense 2DConv in our ablations (Table~2 of the main paper), raising the question of why we implement a native sparse \textit{spatial} rather than \textit{temporal} kernel. We find that three GPU-level constraints make the temporal variant impractical.
\paragraph{Memory locality.}
Tokens within a single frame are stored contiguously: spatial neighbors are at most $\sim$14 positions apart (one grid row), placing the gather within $5$--$11$~KB, well within L1 cache. A temporal neighbor resides in an adjacent frame, $N_{kept/frame} \times C \times 2$ bytes away ($\sim$22--60~KB), an $\sim$8$\times$ increase in access stride. This causes L1/L2 cache thrashing, poor memory coalescing (adjacent warp threads need data from different frame blocks), and higher DRAM traffic as temporal loads cannot be served from cache.
\paragraph{Tiled kernel constraints.}
Our custom tiled kernel gains efficiency from cooperative loading within a single frame block, where spatially nearby tokens share locality. A temporal kernel would require 32 threads to fetch from up to 32 distinct frame blocks simultaneously, causing severe warp divergence and destroying the coalesced access pattern.

\paragraph{Reduced kernel utility under global TopK.}
A sparse temporal kernel would be less \textit{useful} than its spatial counterpart. With global TopK at $k_r{=}0.6$, the probability that a given spatial position $(h,w)$ survives selection in both frames $t$ and $t{+}1$ is approximately $k_r^2 \approx 0.36$. In contrast, spatial neighbors within the same frame share the same selection mask, so a kept token's 8 spatial neighbors have a much higher survival probability, making the spatial kernel's computation substantively meaningful.

In summary, while a native sparse 1D temporal convolution is algorithmically possible, every aspect of its GPU implementation, gather patterns, memory locality, tiling strategy, and effective kernel utilization, is fundamentally worse than the spatial 2D case. The marginal mAP advantage of temporal convolution ($+$0.25\%) does not justify the performance degradation of a native sparse temporal kernel. Our 2D spatial design provides a direct path from token reduction ratios to proportional wall-clock savings.

\end{document}